\documentclass{article}

\usepackage[nonatbib]{neurips_2023}

\usepackage[numbers]{natbib}
\usepackage[utf8]{inputenc} % allow utf-8 input
\usepackage[T1]{fontenc}    % use 8-bit T1 fonts
\usepackage{hyperref}       % hyperlinks
\usepackage{url}            % simple URL typesetting
\usepackage{booktabs}       % professional-quality tables
\usepackage{amsfonts}       % blackboard math symbols
\usepackage{nicefrac}       % compact symbols for 1/2, etc.
\usepackage{microtype}      % microtypography
\usepackage{xcolor}         % colors
\usepackage{graphicx}
\usepackage{amsmath}
\usepackage{booktabs}
\usepackage{pifont}
\usepackage{amsfonts}
\usepackage{amssymb}
\usepackage{dsfont} % indicator function
\usepackage{subcaption}
\usepackage{comment}

\newcommand{\cmark}{\ding{51}}%

\title{Heterogeneous Graph Neural Networks for Short-term State Forecasting in Power Systems across Domains and Time Scales: A Hydroelectric Power Plant Case Study}

\author{Raffael Theiler\\
Intelligent Maintenance and Operations Systems (IMOS)\\
École Polytechnique Fédérale de Lausanne (EPFL)\\
Lausanne, CH-1015, Vaud \\
\texttt{raffael.theiler@epfl.ch} \\
\And
Olga Fink\\
Intelligent Maintenance and Operations Systems (IMOS)\\
École Polytechnique Fédérale de Lausanne (EPFL)\\
Lausanne, CH-1015, Vaud \\
\texttt{olga.fink@epfl.ch} \\
}

\begin{document}
\nolinenumbers

\maketitle

% Abstract
\begin{abstract}%  

Accurate short-term state forecasting is essential  for efficient and stable operation of modern power systems, especially in the context  of increasing variability introduced by  renewable and distributed energy resources.
As these systems evolve rapidly,  it becomes increasingly important to  reliably predict their states in the short term
to ensure operational stability, support control decisions,
and enable  interpretable monitoring of sensor and machine behavior.
Modern power systems often span  multiple physical domains -- including electrical, mechanical, hydraulic, and thermal -- posing significant challenges for modeling  and prediction. 
These challenges arise from complex multi-rate temporal dynamics, 
heterogeneous interdependencies  between subsystems, 
and the presence  of layered control architectures, 
all of which hinder the ability to accurately forecast  short-term system states.
Graph Neural Networks (GNNs) have emerged as a promising data-driven framework 
for system state estimation and state forecasting in such settings. 
By leveraging the topological structure of sensor networks, GNNs can implicitly learn inter-sensor relationships and propagate information across the network.
However, most existing GNN-based methods are designed under the assumption of homogeneous sensor 
relationships and are typically constrained  to a single physical domain -- for example, focusing exclusively on electrical networks.
This limitation restricts  their ability to 
integrate and reason over heterogeneous sensor data commonly encountered in real-world energy systems, such as those used in energy conversion infrastructure.
In this work, we propose  the use  of \textit{Heterogeneous Graph ATtention Networks} to address these limitations. Our approach 
 models both  homogeneous intra-domain and heterogeneous inter-domain relationships among   sensor data from two distinct physical domains -- hydraulic and electrical --  which exhibit fundamentally different temporal dynamics.
By constructing a unified, system-wide heterogeneous graph, that fuses data from these domains, our model is able to capture complex cross-domain interactions that are otherwise neglected in traditional GNN formulations. To address the challenge of   differing sensor sampling rates, 
we introduce a forecasting mechanism  based on first-order forward differentiation,  allowing the model to learn and propagate dynamic behavior more effectively across time. This formulation improves the network’s capacity to  generalize  across temporal scales and enhances predictive accuracy, particularly in ultra-short-term and short-term state forecasting tasks under real-world operating conditions.
We demonstrate the effectiveness of our approach on two real-world datasets from a hydroelectric power plant, comprising sensor measurements sampled at one-minute and one-second resolutions, respectively. Experimental results demonstrate that our method significantly outperforms conventional baselines on average by 35.5\% in terms of normalized root mean square error, confirming its effectiveness in multi-domain, multi-rate power system state forecasting across both case studies
\footnote{Baselines include LSTM, LSTM* (applied signal-wise), StemGNN (el+hyd), 1D-CNN, 1D-CNN*, Persistence in the \textit{1Sec} and \textit{1Min case study}}.
\end{abstract}

\section{Introduction}

The growing integration of distributed energy resources, 
renewable generation, and hybrid power systems has introduced substantial variability into  power system states 
\cite{wuIntegratingModeldrivenDatadriven2022, zhaoPowerSystemDynamic2019}.   
This variability often leads to significant  estimation 
errors when conventional state estimation methods are applied. 
Traditional  approaches typically assume that the current system state is influenced  only by present measurements,
neglecting  the inherent dynamics of power systems
\cite{wangRobustForecastingAidedState2020}.
However, many power system components, such as generators,
demand-side resources,
and energy storage systems,
exhibit dynamic behavior,  responding to both external  inputs
and historical  states.  This creates strong temporal dependencies 
between the current and previous states of the system
\cite{linSpatiotemporalGraphConvolutional2024}.
Consequently, state-space models that explicitly account for  the 
quasi-steady dynamics of the system state and 
integrate temporal correlations of loads and distributed generation 
are  becoming increasingly essential. 
These models enable more accurate tracking of state evolution, 
which is crucial for real-time modeling and control tasks in power systems,
including  forecasting-assisted state estimation,
which integrates the state-transition model into recursive least squares state estimation
\cite{zhaoPowerSystemDynamic2019}
-- a topic that has attracted  growing  research interest in recent years
\cite{
jiRealtimeRobustForecastingaided2021, 
wangRobustForecastingAidedState2020,
zhaoRobustForecastingAided2018a}.
Furthermore, short-term state forecasting
 not only supports state estimation but also facilitates critical  monitoring functions  
such as anomaly and outlier detection for connected assets,
as well as control input validation.
State-space model can also be leveraged to generate
 pseudo measurements,
offering a practical alternative to waiting 
for the next measurement  cycle --
particularly in the presence of frequent meter failures, 
communication delays, 
contingency scenarios, or device malfunction, all of which can result in the loss of real time data 
\cite{alduhaymiForecastingAidedStateEstimation2025, mukherjeeApplicationDeepLearning2021}.
In this context, state forecasting goes beyond conventional state estimation 
by predicting the future evolution of system states 
rather than inferring only their present values. 
Accurate short-term state forecasts provide operators with advance knowledge of impending changes in the grid, 
enabling more proactive and resilient system management.
As a result, state forecasting plays a critical role in maintaining 
the reliable operation and effective control of modern power systems, 
where timely anticipation of system dynamics is increasingly essential
\cite{primadiantoReviewDistributionSystem2017, dehghanpourSurveyStateEstimation2019}.

Modern Energy Management Systems (EMS) play a central 
role to meet the demands of data-driven state-forecasting by incorporating real-time data acquisition, predictive analytics, and system-wide coordination 
\cite{iqtiyaniilhamEuropeanSmartGrid2017, mahmoodImpactsDigitalizationSmart2024}.
EMS data inputs have expanded beyond 
traditional electrical readings to include  time-synchronized measurements 
from multiple physical  domains via  Remote Terminal Units (RTUs) 
\cite{chengSurveyPowerSystem2024}.
For instance, in pumped-storage hydropower plants, hydraulic subsystem variables -- such as
temperatures, lake levels, pressures, and flow rates --
are  collected alongside electrical measurements
\cite{theilerGraphNeuralNetworks2024}
Because energy conversion involves complex interactions across electrical, 
mechanical, and thermal domains,  
these heterogeneous sensors capture cross-domain dynamics 
that are not observable through a single modality.
Consequently, modern power system behavior must be understood 
in terms of both intra-domain homogeneity and inter-domain heterogeneity, both of which evolve over time.
Traditional modeling approaches, however, often assume inter-system homogeneity, or impose narrow system boundaries, 
inadequate for accurate short-term forecasting in dynamic, hybrid systems 
\cite{zhaoPowerSystemDynamic2019}.

These limitations are further compounded by the fact that the underlying physical processes 
operate on distinct and often widely varying time scales.
Electrical variables -- such as current, active power, 
and reactive power -- can exhibit abrupt changes  from one scan to the next, 
particularly  during switching events or rapid load variations.
In contrast, hydraulic and thermal variables tend to evolve more  gradually due to the inertia of mechanical components and fluid dynamics. 
This disparity introduces  a significant modeling challenge 
\cite{yiMstiGnnMultiScaleSpatiotemporal2025}. 
Conventional methods that focus on slow trends -- such as Kalman smoothers with low process noise or autoregressive models 
designed for long-term drift -- often overlook fast electrical transitions
\cite{zhaoRolesDynamicState2021}.
Conversely, models optimized  for rapid signal changes, 
such as temporal convolutional networks with short receptive fields 
or high-order autoregressive models, may fail to capture the slower cross-domain dynamics 
that shape  the evolving system state. This misalignment between model focus and system dynamics  can result in inconsistent
 predictions and increased risk of overfitting.
Developing a unified forecasting approach that remains accurate across domains  
with heterogeneous  sampling rates and dominant time scales remains  an open problem. 
Such a model must effectively capture both fast electrical responses 
and slower physical dynamics, without degradation  in performance
when deployed across diverse sensor configurations and asset types
\cite{zhaoRolesDynamicState2021, chengSurveyPowerSystem2024, kamyabiComprehensiveReviewHybrid2024}.

Although the growing availability of heterogeneous data sources 
 has improved  observability across physical domains in power systems, 
it has also significantly increased  modeling complexity.
Traditional  state estimation algorithms often struggle  to converge under sparse sensing conditions  and
are not well-suited for explicitly modeling multi-domain interactions 
governed by physical laws
\cite{habibDeepStatisticalSolver2023, zamzamDataDrivenLearningBasedOptimization2019}.
The slow convergence of conventional algorithms often results in outdated forecasts and rapid divergence between actual and estimated system  states. This discrepancy can trigger false positives during real-time anomaly detection and hinder the timely identification of critical failures\cite{sweeneyFutureForecastingRenewable2020}.
To address  these limitations, a variety  of data-driven approaches have been proposed 
to model  system behavior without relying solely   on first-principles
\cite{kundacinaStateEstimationElectric2022, jinNewTrendState2021}.

In  light of the limitations of traditional physics-based models and purely data-driven approaches, graph learning has emerged as a state-of-the-art  paradigm for modeling power system dynamics
\cite{liaoReviewGraphNeural2022}. 
By incorporating both
the topological structure of the power grid and   time-series sensor  data,  
graph-based models are well-suited to capture the complex relational dependencies inherent in these  systems.
Among them, Graph Neural Networks (GNNs) have demonstrated  strong performance 
in state estimation by leveraging the  sensor network's structure. However, 
most existing GNNs   assume homogeneous  and uniform relationships between  nodes, which limits their ability to model  the true complexity and heterogeneity of real-world systems.

To address   these shortcomings, 
we propose the use of Heterogeneous Graph ATtention Networks (HGATs) 
to model the distinct energy domains and interaction types present in energy conversion and hybrid energy systems. This approach enables end-to-end learning of system dynamics  
while explicitly accounting for interactions with control signals
\cite{ilicHierarchicalPowerSystems1996}.
Our implementation performs short-term state forecasting using a data-driven HGAT that functions as a discrete-time state-transition model.  It predicts first-order forward finite differences in state variables based on latent time encodings of previous sensor readings, 
thereby improving the accuracy under quasi-steady-state conditions.
This time-then-graph paradigm has been shown to be a powerful framework for homogeneous spatio-temporal signal processing
\cite{gaoEquivalenceTemporalStatic2023}.
Compared to homogeneous graph models, our method achieves higher predictive accuracy, 
and unlike conventional  numerical simulations, it is computationally efficient 
and does not require expert-driven parameter calibration. The explicit encoding  of heterogeneous graph structures also improves interpretability of learned representations.
Furthermore, the proposed approach is  highly  transferable across assets, without the need for system-specific customization. With GPU-accelerated inference,  it supports real-time processing of incoming  sensor data, making it well-suited for deployment in dynamic and operationally diverse industrial environments.

We validate our proposed framework using two real-world case studies from a pumped-storage hydropower plant (PSH) -- a representative example of renewable energy conversion systems that tightly couple two distinct physical domains: hydraulic and electrical.
In modern power grids, PSH plants are a well-established solution 
for large-scale energy storage, valued for their efficiency, scalability and operational flexibility.
They play a key role in balancing supply and demand by dynamically responding to  fluctuations in grid load through controlled  management 
of water  reservoirs \cite{santosPiecewiseLinearApproximations2022}.

This work presents  a heterogeneous graph attention  network framework  designed to  learn both intra-domain and inter-domain dependencies among sensors in power systems 
 -- specifically demonstrated on a PSH system --  
with the goal  of improving  short-term state forecasting. 
Our approach addresses several key challenges posed by the complex and dynamic behavior of PSH operations:

\begin{itemize}
    \item \textbf{Capturing complex sensor dynamics:} PSH systems  operate across multiple modes (e.g., pumping and generating), leading to complex, time-varying sensor behavior. Our HGAT model  effectively captures these dynamics  by incorporating structural priors such as connectivity,
    sensor placement, and recurring operational  patterns. These priors support  relation-aware homophily in the learned graph structure, enabling more accurate representation of sensor interactions.
    \item \textbf{Learning from exogenous and unmodeled factors:}   Factors such as ambient temperature,  grid-level demand fluctuations, 
    and consumer behavior significantly influence PSH system states  but  are typically not represented in physics-based simulations.
    Our model leverages spatio-temporal graph learning to extract these  dependencies directly from data, improving predictive performance under realistic conditions.
    \item \textbf{Modeling heterogeneous physical subsystems:} PSH plants   integrate  hydraulic and  electrical  subsystems, each governed by different  physics and sensing modalities.
    We address this complexity by constructing a heterogeneous graph that encodes both within-domain and cross-domain relationships, enabling the model to account for multi-physics interactions.
    \item \textbf{Time-then-graph Paradigm:} We demonstrate  that the time-then-graph modeling strategy is effective  across time scales in power systems datasets, supporting robust performance in both fast and slow process regimes.
    \item \textbf{Comprehensive validation across time scales:} We  evaluate the proposed HGAT model on two multivariate datasets from a real-world PSH installation. The datasets include 121 and 46 synchronized sensor signals at minute and second-level resolutions, respectively. In contrast to prior work with limited  feature sets or domain scope,
    our approach leverages RTU measurements, flow sensors, and other  domain-specific signals, such as lake levels -- to deliver  more precise and robust state forecasts.
\end{itemize}

The remainder of this paper is organized as follows: 
Sec. \ref{sec:background_and_related_work} provides an overview of related work in graph-based deep learning and multi-domain  data fusion. Sec. \ref{sec:methodology}
introduces the proposed  heterogeneous graph neural network approach. 
 Sec.~\ref{sec:case_study} presents  the case study conducted on a Swiss pumped-storage hydropower plant, detailing  the experimental design  and training setup.  Sec.~\ref{sec:results} presents  the  results and evaluates the model’s performance. 
Finally, Sec.~\ref{sec:conclusions} summarizes the key  findings and outlines  directions for future research.

\section{Background and Related Work}
\label{sec:background_and_related_work}

Traditional approaches to power system state estimation
rely on techniques such as exponential smoothing,
recursive least squares,
and Kalman filtering
\cite{zhaoPowerSystemDynamic2019, kamyabiComprehensiveReviewHybrid2024}.
While these methods are computationally efficient and analytically  well-founded, 
they often struggle to  produce accurate estimates  
under  abrupt changes in system state -- such as  sudden load variations, 
fluctuations in distributed energy resources, or network topology reconfigurations.
These  shortcomings stem primarily from to their reliance on linear state transition models and their limited  adaptability to dynamic system behaviors
\cite{zhaoPowerSystemDynamic2019}.
To address   these limitations, 
machine learning models capable of  capturing non-linear 
system dynamics have gained traction in recent years. 
Early approaches    primarily employed  
recurrent neural networks (RNNs)  and convolutional neural networks (CNNs)
~\cite{zhengElectricLoadForecasting2017}, 
which have demonstrated strong performance in tasks  such as  
short-term load forecasting
~\cite{guoMachineLearningBasedMethods2021}
and peak electricity  demand prediction 
\cite{kimDailyPeakElectricityDemandForecasting2022}.
However, these approaches typically process all input features in a uniform manner, without explicitly modeling the distinct relationships and mutual influences encoded in the system’s topology. 
As a result, their applicability to hybrid systems is limited, where such structural dependencies critically shape state evolution.

\paragraph{Graph Neural Networks}
With the growing recognition of the graph-structured nature of power grids, Graph Neural Networks (GNNs)
-- originally introduced by Bronstein et al.~\cite{bronsteinGeometricDeepLearning2017} -- 
have emerged as a state-of-the-art method for modeling power system behavior.
GNNs offer the ability to incorporate  topological relationships and encode structural biases that traditional models often overlook.
Several GNN-based approaches have been proposed for state estimation, particularly those built on \textit{Graph Convolutional Neural Networks} (GCNs) 
\cite{fatahIntegratingPowerGrid2021}, 
often leveraging   DC power flow approximations  
\cite{wuIntegratingModeldrivenDatadriven2022}.  
These have been extended  through more expressive architectures, such as graph Transformers
\cite{ringsquandlPowerRelationalInductive2021} 
and  physics-informed GNNs that integrate domain knowledge into the learning process \cite{pagnierPhysicsInformedGraphicalNeural2021}. 
GNNs have also  been used for  quasi-steady-state estimation  
\cite{kundacinaStateEstimationElectric2022}
and for high-frequency  state estimation using 
Phasor Measurement Unit (PMU) data 
\cite{linSpatiotemporalGraphConvolutional2024}
.
Beyond estimation, GNNs have demonstrated effectiveness in a variety of  downstream tasks across power systems, including
anomaly detection in smart grids 
\cite{liDynamicGraphBasedAnomaly2022} 
and dynamic grid stability prediction 
\cite{nauckPredictingBasinStability2022}. 
Despite these advancements, 
most existing GNN frameworks are limited by their assumption of homogeneous node interactions 
and their focus on single-domain data.  As such, they are not well-suited to multi-domain systems, where
interactions between electrical, hydraulic, and thermal subsystems must be modeled as distinct, domain-specific functions. Addressing this limitation is critical for improving the robustness and generalizability of GNN-based methods in complex, hybrid energy systems.

\paragraph{Heterogeneous Graph Neural Networks}
Recent research has increasingly focused on  heterogeneous graph neural networks (HGNNs),  which extend traditional GNNs by enabling  node- and edge-type-specific interactions. Originally  introduced for heterogeneous graph classification tasks \cite{zhangHeterogeneousGraphNeural2019}, HGNNs have gained traction 
in power system applications, where diverse physical entities and their interactions cannot be adequately captured by homogeneous graph models  \cite{liaoReviewGraphNeural2022}.
In power systems, HGNNs  have been applied  to a range of tasks, including  multi-region forecasting  of wind and solar  power generation   \cite{liHeterogeneousSpatiotemporalGraph2024}, modeling  electrical bus systems for optimal power flow analysis \cite{ghamiziOPFHGNNGeneralizableHeterogeneous2024}, and short-term voltage stability  assessment  \cite{lvShortTermVoltageStability2025}.
Beyond pairwise interactions, hypergraph extensions have been proposed for distribution systems to 
enable more expressive representations 
through  node-centric message-passing
\cite{habibDeepStatisticalSolver2023}. 
While these approaches offer enhanced modeling capacity, existing HGNNs
have not yet been applied to fuse information across  multiple physical domains 
for the task of short-term power system state forecasting. This gap highlights the need for models that can capture both domain-specific dynamics and cross-domain dependencies in a unified, data-driven framework.

% data fusion
\paragraph{Data Fusion}
An important area of recent advancement  in power systems is  data fusion, which improves forecasting  accuracy and system observability by integrating information from diverse  sensing modalities.
While  Supervisory Control and Data Acquisition (SCADA) systems remain the foundation  of traditional data collection, the growing  deployment of
Advanced Metering Infrastructure (AMI), 
Intelligent Electronic Devices (IEDs), 
and
Phasor Measurement Units (PMUs)  has significantly    broadened the range and resolution of data available for state estimation 
\cite{chengSurveyPowerSystem2024, kamyabiComprehensiveReviewHybrid2024}.
 Modern power systems increasingly incorporate additional sensor types -- such as transformer tap positions, 
line temperature sensors, 
and thermal and hydraulic measurements -- particularly in the context of integrated energy systems \cite{chengSurveyPowerSystem2024}.
To leverage these heterogeneous data sources, various data fusion  strategies have been developed. For instance, voltage estimation  in distribution networks  has been improved by exploiting cross-correlations among  transformer measurements
\cite{zhuCrossDomainDataFusion2020}.
Similarly, multi-site photovoltaic (PV) power forecasting has benefited from spatially distributed PV data,  which effectively function  as a dense network of virtual weather stations \cite{simeunovicSpatioTemporalGraphNeural2022}.
In hydropower plants, prior work has demonstrated that fusing sensor data from both hydraulic and electrical subsystems -- which are typically governed  by a shared control input -- can significantly improve the accuracy of state forecasting \cite{theilerGraphNeuralNetworks2024}. However, existing methods have not yet leveraged  heterogeneous graph neural networks for this task. As a result, all cross-domain interactions are modeled using shared weights,   limiting the ability to capture domain-specific dynamics and transfer functions.

To address  this gap, we   propose a novel heterogeneous  graph-based framework  that, for the first time, explicitly models cross-domain interactions between hydraulic and electrical subsystems in power plants. Our approach integrates self-attention mechanisms, domain-specific edge types, and structural priors derived from system schematics, while also capturing  temporal dependencies across multiple time scales. This unified formulation enables accurate short-term state forecasting by learning both intra- and inter-domain dynamics, setting it apart from existing models that treat all signals uniformly or overlook cross-domain coupling.

\section{Methodology}

{\small
\textbf{Notation}: In this work, we use slicing notation denoted by the colon symbol (\(:\)).
Given a matrix \( A \in \mathbb{R}^{m \times n} \), where \( m \) and \( n \) denote   the number of rows and columns respectively, a slice is expressed as \(A[i:j, k:l]\) or \(A^{i:j, k:l}\).
This notation represents the selection of rows \( i \) through \( j-1 \) and columns \( k \) through \( l-1 \) of matrix \( A \). Omitting   \( i \) or \( k \)  implies  selection   from the first row or column, while  omitting   \( j \) or \( l \)  implies selection up to  last row or column.
We use $\otimes$ to denote element-wise multiplication,  $\oplus$ for concatenation, and \(\Vert \bullet \Vert_{\text{F}}\)  for the Frobenius norm. The symbol  $\times$ denotes   the Cartesian product.

}

\begin{figure*}[tph]
    \centering
    %\fbox{\rule[-.5cm]{0cm}{5cm} \rule[-.5cm]{5cm}{0cm}}
    \includegraphics[width=1\linewidth]{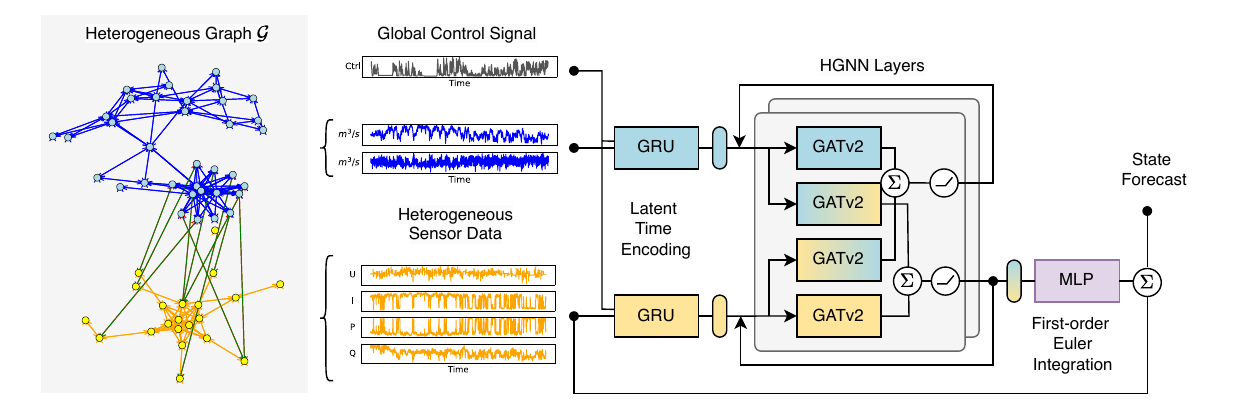}

  \caption{
  An overview of the processing steps of our proposed \textit{Heterogeneous Graph ATtention Network} (HGAT), applied to the case study of a pumped-storage hydropower plant. By operating on a heterogeneous graph, the HGAT efficiently extracts information from the hydraulic and electrical sensor data to forecast the electrical state variables. }
    \label{fig:methodology_overview}
\end{figure*}

\label{sec:methodology}
This chapter presents our proposed framework based on  a \textit{Heterogeneous Graph ATtention Network} (HGAT), designed   to model the complex relational dependencies present  in multi-modal time series data collected from power systems. 
The HGAT architecture learns relation-specific   message-passing functions across node and edge type pairs, enabling context-aware forecasting of system dynamics by leveraging  attention mechanisms over sensor and control data alone.
The key novelty of our method lies  in  the integration   of structural priors --derived   from system schematics -- with attention-based message passing over   a heterogeneous graph. 
In addition, we reformulate the forecasting problem    as one of  derivative prediction with forward integration, which  allows the model to   capture system dynamics more robustly across datasets with varying   temporal resolutions .

As a core contribution, we propose a  modular architecture comprising  three complementary components, designed to work either independently or in combination for enhanced state forecasting performance.
The proposed method  begins by augmenting the system state with first-order forward differences, explicitly   capturing short-term dynamics. These enriched representations are then passed  through  a gated recurrent unit (GRU)-based encoder, which models  temporal dependencies and generates  latent  states for each sensor node. 
To incorporate spatial and relational context, the latent system state representations are refined in the subsequent  module through  attention-based message passing over a heterogeneous graph, enabling the model to account for both intra- and inter-modality interactions. 
Finally, the model predicts  finite differences from the refined node embeddings,   which are then integrated using Euler's method  to estimate the next  system state --  resulting in a forecasting approach that is both dynamically informed and guided by structural priors. 
Each  step   is described  in detail in the following sections.

\paragraph{First-order Forward Differentiation and System State}
\label{sec:method_timeseries_data}
Let the sensor and control signal be defined  matrices as follows :
\begin{itemize}
    \item \( \mathbf{X} \in \mathbb{R}^{T \times D} \): Time series of sensor measurements over  \( T \)  time steps and \( D \)  sensor channels. 

    \item \( \mathbf{U} \in \mathbb{R}^{T \times K} \): Time series of control inputs over  \( T \)  time steps and \( K \) control variables.
\end{itemize}
To explicitly capture short-term temporal dynamics, we compute the  first-order forward finite differences of both sensor and control signals:  

\[
\dot{x}(t) :=\frac{dx(t)}{dt} \approx \frac{x(t+1) - x(t)}{\Delta t},
\]  

where \( \Delta t \) denotes the discrete time step ,   set to match  the sampling interval   of the dataset. The resulting temporal derivatives \( \dot{\mathbf{X}} \) and \( \dot{\mathbf{U}} \) are concatenated   with the original signals to form the model input matrix:  \( \mathbf{S} = (\mathbf{X},\dot{\mathbf{X}},\mathbf{U},\dot{\mathbf{U}}) \), which enriches the representation with both absolute and differential features, providing a more informative state description for subsequent learning modules.

\paragraph{State Forecasting by Euler Integration} 
\label{para:state_forecating_model}
We define the  forecasting model as a function:
\(
M: \mathbb{R}^{w \times (2D+2K)} \rightarrow \mathbb{R}^{h \times d}
\)
where \( w \) denotes the size of  the input time    window,  
 \( h \) is the prediction horizon, and \( d \leq D \) is the number of target output  variables.  The input includes both original and first-order differential features of sensor and control data.

The model  outputs a sequence of predicted  temporal derivatives :
\(
\dot{\mathbf{X}}[t:t+h, :d],
\)
over a forecast horizon of length \( h \).  These derivatives are then integrated using Euler integration  to reconstruct the forecasted signal values  :
\begin{equation}
\label{eq:forward_integration}
\widetilde{\mathbf{X}}[t+k, :d] = \mathbf{X}[t-1, :d] + \sum_{i=0}^{k} \widetilde{\dot{\mathbf{X}}}[t-1+i, :d], \quad \text{for } k = 0, \ldots, h-1.
\end{equation}
All operations are performed within a sliding window of size $w$, restricting the model’s access to past data and enforcing   a real-time forecasting   constraint. This setup enables dynamic-aware prediction while maintaining operational feasibility.

\paragraph{Node Representations}
To connect  time-evolving system states \( \mathbf{S} \) to the heterogeneous graph structure   \( \mathcal{G} \), we assign each node \( v \in V \) a dedicated slice   of the input data, denoted \( \mathbf{S}_v \subseteq \mathbf{S} \). These node-specific inputs  are disjoint subsets   that capture local  observations:
\[
\mathbf{S}_v[t-w:t] \in \mathbb{R}^{w \times n}, \quad n < (2D + 2K)
\]
Each node can also  incorporate additional  covariates, such as time encodings or global control signals . 
To embed these inputs into a common representation space, we define a type-specific   embedding function:
\[
\text{Embed}_{\phi(v)}: \mathbb{R}^{w \times n} \rightarrow \mathbb{R}^{d_{emb}}
\]
which produces the initial hidden state for each   node:
\[
\mathbf{h}_v^{0} = \text{Embed}_{\phi(v)}(\mathbf{S}_v[t-w:t]), \quad \forall v \in V
\]

\paragraph{Heterogeneous Message Passing}
The proposed \textit{Heterogeneous Graph ATtention Network} (HGAT)
 operates on a heterogeneous graph defined as: \(
\mathcal{G} = (V, E, \mathcal{T}_V, \mathcal{T}_E)
\)
where \( \mathbf{
V
} \) and \( \mathbf{
E
} \) are the sets of nodes and edges, and  \( \mathcal{T}_V \)  and 
\( \mathcal{T}_E \) denote the sets of node and edge types, respectively.
To model the diverse interactions present in multi-modal sensor networks, HGAT defines a distinct message-passing function for each edge type \( t \in \mathcal{T}_E \). These type-specific  functions govern how information flows between nodes, accounting for both  the structural and semantic heterogeneity of the system .
The general form of a message function for edge type \(t\) is given by :

\[
M_{t}(h_u, h_v, e_{uv}) = f_{\theta_t}(h_u, h_v, e_{uv})
\]

where \( M_t \) is the message function for edge type \( t \in \mathcal{T}_E \),
\( h_u\) and \( h_v \) are the current  latent  representations of the source node \( u \) and target node  \( v \), respectively, and   
\( e_{uv} \) denotes   optional edge features.
The function \( f_{\theta_t} \), parameterized by \( \theta_t \), is a learnable neural network  specific to edge type \( t \).
The initial node embeddings \(h_\bullet^ {l=0}\) for the input layer \(l=0\) are derived   from the sensor input window \( \mathbf{S}_v[t-w:t] \), using a type-specific encoder: \[
h_v^{(0)} =\text{Embed}_{\phi(v)}(\mathbf{S}[t-w:t]) \]This embedding includes  raw sensor data, control inputs , and their   first-order forward differences,  enabling the model to encode both current system states and short-term dynamics .
By assigning separate attention-based message-passing functions to each edge type, HGAT effectively captures the heterogeneous interactions induced by the physical topology and operational interdependencies of the system. This allows the model to adapt its aggregation strategy to reflect domain-specific relationships among components, improving its ability to learn from complex, multi-domain data.

\paragraph{Graph Attention Operator}
The input node representations \(\mathbf{h}_v^{0}\) are iteratively refined  through $L$ layers of heterogeneous graph attention. 
At each layer \( l \), the  representation of node \( v \) is updated by aggregating messages from its neighbors \( v_n \in \mathcal{N}_t(v) \), where each message is computed by a type-specific function  \(M_{t}\)
 for edge type \( t \in \mathcal{T}_E \). We use  summation as the aggregation operator and apply a non-linear activation function \( \sigma \) to the aggregated result:

\begin{equation}
h_v^{(l+1)} = \sigma \left( \sum_{t \in \mathcal{T}} \sum_{v_n \in \mathcal{N}_t(v)} M_t \left( h_v^{(l)}, h_{v_n}^{(l)}, e_{v_n v}^{(t)} \right) \right)
\end{equation}

This formulation enables the model to learn both intra- and inter-type interactions, capturing the diverse relational semantics present in heterogeneous graphs.

For the message passing operator \(M_{t}\), we adopt  Graph Attention Networks (GATv2) \cite{brodyHowAttentiveAre2022}, which enable adaptive, data-driven aggregation of information  from neighboring nodes. Unlike static  aggregation schemes, GATv2  assigns learnable  weights to incoming messages, allowing the model to emphasize  the most informative  interactions. 
This is particularly beneficial in heterogeneous graphs, 
where edge types represent   semantically  distinct relationships that influence node representations  differently. The message passing  function for an edge of type $t$ is defined as:

\[
M_t \left( h_v^{(l)}, h_{v_n}^{(l)}, e_{v_n v}^{(t)} \right) = \alpha_{vu} \mathbf{W}_{\text{gat}}^{(l)} \mathbf{h}_u^{(l)}
\]

where \( \mathbf{W}_{\text{gat}}^{(l)} \) is a learnable weight matrix at layer \( l \), and $\alpha_{vu}$ is the attention coefficient  that quantifies the relative importance of node $u$'s message  to node $v$, conditioned on the edge type \( t \).

The attention coefficient 
$\alpha_{vu}$
 is computed as:

\[
\alpha_{vu} = \frac{
    \exp\left( \text{LeakyReLU}\left( \mathbf{a}^\top \cdot \mathbf{W}_a \left[ \mathbf{h}_v^{(l)} \| \mathbf{h}_u^{(l)} \right] \right) \right)
}{
    \sum_{u' \in \mathcal{N}_{\text{het}}(v)} 
    \exp\left( \text{LeakyReLU}\left( \mathbf{a}^\top \cdot \mathbf{W}_a \left[ \mathbf{h}_v^{(l)} \| \mathbf{h}_{u'}^{(l)} \right] \right) \right)
}
\]

where $\mathbf{a}$ is a learnable attention vector, \( \mathbf{W}_a \in \mathbb{R}^{d \times 2d_h} \) is a learnable weight matrix applied to the concatenated node embeddings, and $\|$ denotes  concatenation.
After $L$ layers of heterogeneous attention-based message passing , the final node representations are given by:
\[
\mathbf{z}_v = \mathbf{h}_v^{(L)}, \quad \forall v \in V.
\]

\paragraph{Graph Head}
The final  graph representations \(\mathbf{z}_v\) encode structural and contextual information relevant for predicting first-order forward differences  at each sensor node $v$.  
To project  these latent embeddings back into the signal space, we apply  a two-layer feedforward network -- referred to as the \textit{graph head} -- comprising the sequence:  Linear \(\rightarrow\) LeakyReLU \(\rightarrow\) Linear. This head outputs  the estimated first-order forward differences  \(\widetilde{\dot{\mathbf{X}}}_t\) for each node:

\[
\widetilde{\dot{\mathbf{X}}}_t = \mathbf{W}_2 \cdot \text{LeakyReLU}(\mathbf{W}_1 \cdot \mathbf{z}_v + \mathbf{b}_1) + \mathbf{b}_2
\]

These  estimated derivatives are then recursively integrated using the Euler method described in Equation Equation \ref{eq:forward_integration} to obtain the predicted sensor signal trajectories.

\paragraph{Training Objective}
\label{sec:method_training_objective}

Using  the sliding window scheme  described  in Section \ref{para:state_forecating_model}, we construct the training dataset $\mathcal{X}^{\text{Train}}$ of length \(N_\text{train}\) based on the selected  training split.
Analogously, we define the validation and test datasets as $\mathcal{X}^{\text{Val}}$ and $\mathcal{X}^{\text{Test}}$ , respectively:
\[\mathcal{X}^{\text{Train}} =  \{ \mathbf{S}^{t:t+h}, \mathbf{S}^{t-w:t} \}_{t=w}^{N_\text{train}}
\]

The model  is trained to minimize the discrepancy   between the predicted and ground-truth first-order forward differences of the target sensor signals using the following loss function:
\begin{equation*}
    \mathcal{L}(\mathbf{\widetilde{\dot{X}}}, \mathbf{\dot{X}}) = \sum_{t=w}^{N_\text{train}} \| \widetilde{\mathbf{\dot{X}}}^{t:t+h} - \widetilde{\mathbf{\dot{X}}}^{t:t+h} \|_F^2
\end{equation*}
We then determine the optimal model parameters  by solving the following optimization problem:
\begin{equation}
    \theta_t^*  = \arg\min_{\theta_t, \phi} \mathcal{L}(\mathbf{\widetilde{\dot{X}}}, \mathbf{\dot{X}}; \theta_t)
\end{equation}

Here, \(\theta\) and \(\phi\) collectively represent the learnable parameters of the message-passing functions and node encoders, respectively.

\section{Case Studies and Experimental Setup}
\label{sec:case_study}

\subsection{Data Collection}
The datasets used in this case study were collected in collaboration with the Swiss Federal Railways (SBB), 
which operates a dedicated railway traction network
running at 16.7 Hz to supply electricity to rolling stock across Switzerland.
Sensor data is transmitted via \textit{Supervisory Control and Data Acquisition (SCADA)} protocols 
to a centralized \textit{Energy Management System} (EMS). 
Although  SCADA data is often not inherently time-synchronized, SBB's  event-based 
wide-area measurement infrastructure ensures precise temporal alignment across  sensor streams.
This setup  supports  real-time
asset monitoring 
and provides a   technically  robust environment   for evaluating the proposed methodology, particularly in terms of maintaining temporal consistency across heterogeneous sensor data.

\begin{figure*}[tph]
    \centering
    %\fbox{\rule[-.5cm]{0cm}{5cm} \rule[-.5cm]{5cm}{0cm}}
    \includegraphics[width=1\linewidth]{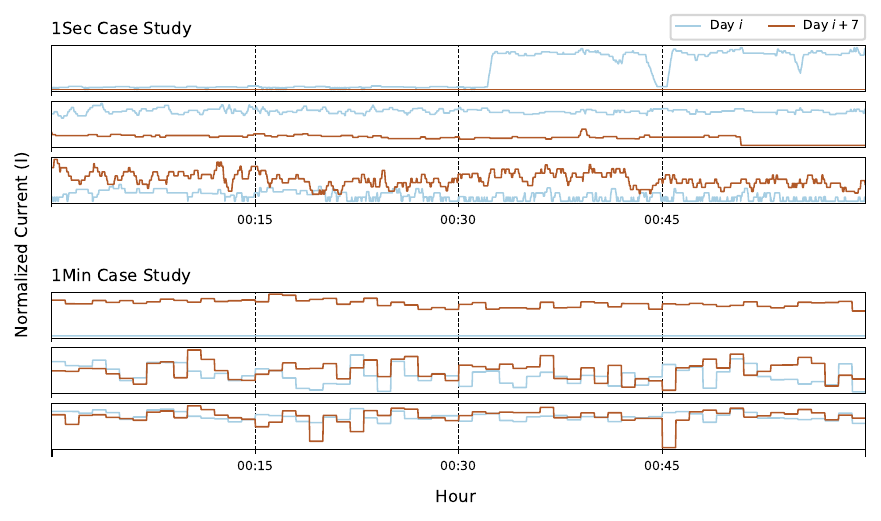}

  \caption{
 Example segment from the \textit{1Sec case study} and \textit{1Min case study}, illustrating three time-synchronized normalized current measurements (randomly chosen from the data sets for demonstration) sampled at different temporal resolutions. The plots correspond to the same hour  on two consecutive weeks (Day $i$ and Day $i+7$), highlighting both the dynamic behavior of the signals and characteristic weekly patterns.
 }
    \label{fig:time_series_overview}
\end{figure*}

An example segment from the 1Sec and 1Min case studies  is shown in Figure \ref{fig:time_series_overview}, illustrating normalized current measurements at different temporal resolutions. Each plot shows the same hour on two consecutive weeks (Day i and Day i + 7), highlighting both the dynamic variability of the signals and recurring weekly patterns.

\textbf{Objective:} For this case study, 
we focus  on a single energy conversion system
-- a pumped-storage hydropower plant -- 
connected to the railway traction power network.
The objective is to forecast the state 
of the electric subsystem, which is
monitored using  high-resolution measurements from Remote Terminal Units (RTUs).  
These measurements include active power (P), 
reactive power (Q), 
voltage magnitude (U), 
and current magnitude (I),
recorded  at 1Hz or \( \frac{1}{60}~\mathrm{Hz} \)  sampling frequency.
We refer to this dataset as \textit{RTU data}.
From an operational standpoint, 
forecasting these RTU measurements 
is particularly relevant  for railway systems due to the highly dynamic and non-residential load profiles  
of rolling stock.
Unlike residential power grids, where load fluctuations are comparatively moderate, the railway traction network (RTN) operated by SBB is subject to far more abrupt changes. For instance, while  Zurich  residential grid -- the largest City of Switzerland -- can experience transient load changes  of up to 35MW within 15-minute intervals,
the RTN regularly undergoes fluctuations of up to 250 MW over  the same duration.   These sharp variations are driven by the coordinated and periodic timetable of the national rail system, also referred to as "regular interval timetable" or the "Swiss integrated timetable" \cite{halderPowerDemandManagement2018}.

\textbf{Datasets \& Data Preparation:} 
We collected two datasets for this case study: one with  one-second resolution and another with one-minute resolution.
 The \textbf{one-minute resolution dataset} (referred to as the \textit{1Min case study}) captures  four months  of operation (January to March 2021) at  a PSH facility in Switzerland. It 
consists of measurements  from 37 sensors monitoring  the hydraulic subsystem, such as pressures, flow rates, and lake levels, as well as RTU data 
from seven generating units and their associated substations 
in the electrical subsystem. In total, the dataset comprises measurements from 84 sensors.

\begin{figure}[ht]
    \centering
    \begin{subfigure}{0.45\textwidth}
        \centering
        \includegraphics[width=\linewidth]{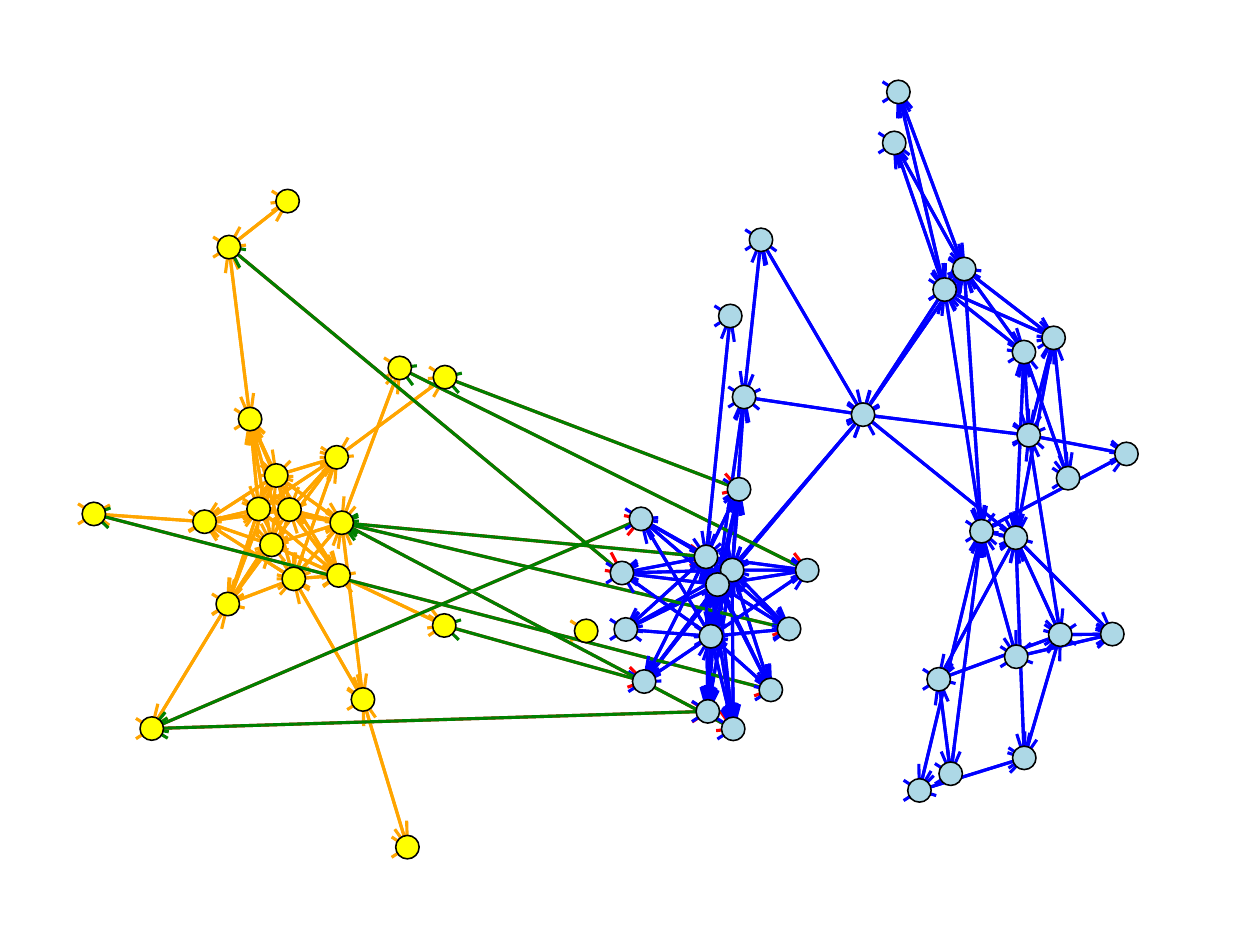}
        \caption{Heterogeneous graph of the \textit{1Min case study}.}
        \label{fig:left}
    \end{subfigure}
    \hfill
    \begin{subfigure}{0.45\textwidth}
        \centering
        \includegraphics[width=\linewidth]{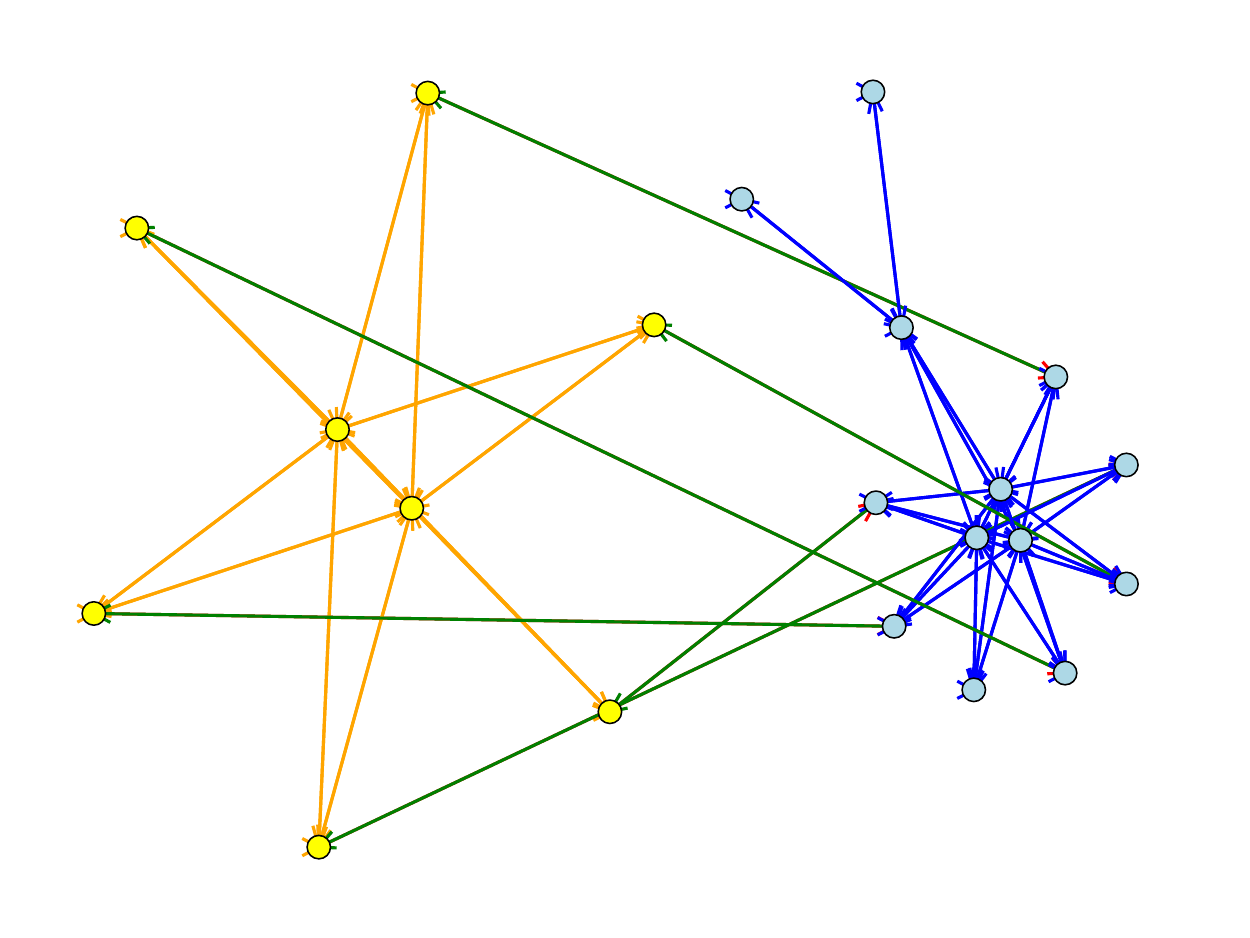}
        \caption{Heterogeneous graph of the \textit{1Sec case study}.}
        \label{fig:right}
    \end{subfigure}
  \caption{The heterogeneous graphs of the pumped-storage hydropower plant case studies.}
      \label{fig:graphs}
\end{figure}

To ensure   temporal consistency, we split  the dataset chronologically into training (70\% ), validation (15\%) and test (15\%) sets,  with  validation and test periods  strictly following  the training interval. 
Feature-wise min-max scaling is applied for normalization, and first-order forward differences of the system states are standardized using z-scoring.
A representative  segment of this dataset is illustrated in Figure~\ref{fig:time_series_overview}, highlighting  its temporal dynamics.
The \textbf{one-second resolution dataset } (referred to as the\textit{1Sec case study}) ,
spans selected days between April and March 2023, comprising 345,000 training samples.
Due to  system security constraints, 
exporting large volumes of  high-frequency data was not feasible.
As a result, this dataset includes  measurements from 8 RTUs, and 13 sensors monitoring the hydraulic subsystem, collected  over several days selected based on dynamic load conditions.
Figure~\ref{fig:graphs} illustrates the heterogeneous graph structures of both datasets.

\subsection{Representing the PSH as HGAT}
\label{sec:case_study_theory}

In hybrid energy systems such as pumped-storage hydropower (PSH), multiple physical subsystems interact under shared operational control. These systems generate  time-synchronized, multimodal  data across domains 
 (e.g., electrical and hydraulic), and their component layout is naturally represented as a graph.  To demonstrate the practical utility of our proposed  framework (introduced in Section \ref{sec:methodology}), we apply it 
to the pumped-storage hydropower (PSH)  case study introduced in Section \ref{sec:case_study}.

\paragraph{Heterogeneous Graph}
For the PSH system, we define the set of node types as \( \mathcal{T}_V = \{\text{elec}, \text{hydro} \} \), corresponding to the electrical and hydraulic subsystems. The set of edge types  are defined as \( \mathcal{T}_E = \{\text{elec-elec}, \text{hydro-hydro}, \text{elec-hydro}, \text{hydro-elec} \} \), capturing both intra- and inter-domain interactions.
The heterogeneous graph $\mathcal{G}_\phi = (V, E)$ is is derived from detailed schematic diagrams  of the electrical and hydraulic subsystems of the PSH plant. Nodes \(v \in V\) represent physical  components or sensor locations  such as generators, turbines, penstocks, and reservoirs. 
In the case of the \textit{1Sec case study}  nodes also include components from the connected substations. 
Each node is uniquely categorized as either electrical or hydraulic:
\(
\mathds{1}_{\text{el}}(v) = 1\) for electrical components, and \( \mathds{1}_{\text{hyd}}(v) = 1\) for hydraulic components.
ensuring  mutual exclusivity of domain membership.
The set of edges \(E\) is defined according to  the physical  interconnections specified   in the system  schematics:

\begin{itemize}
    \item In the electrical subsystem,  edges are defined by physical connections between components such as transformers, switchgear, and generators.
    \item In the hydraulic subsystem, edges correspond to connections between penstocks, pipelines, valves, and turbines.
    \item  Heterogeneous edges represent the mechanical drive coupling between  hydroelectric turbines 
    and electromagnetic generators, linking the hydraulic and electrical domains, modeling the physical link through which energy conversion occurs.
\end{itemize}

Sensors are organized into nodes, with each node comprising $k > 0$ sensors that capture co-located or functionally related measurements.
For example, RTU nodes of the electrical subsystem typically measure  voltage (U), current (I), and compute active (P) and reactive power (Q), resulting in $k=4$ features. Nodes thus represent groups of sensor signals localized to a physical site and categorized by subsystem.
This graph structure  serves as the backbone of   the HGAT,  enabling structured message passing across subsystem boundaries and facilitating the learning of cross-domain dependencies critical for accurate state forecasting.

\paragraph{Electrical State Forecasting in PSH Systems}

The objective of this study  is to perform short-term state forecasting by
predicting first-order forward finite differences  in sensor signals  from  the electrical subsystem of a PSH, using both electrical and hydraulical information under quasi-steady-state conditions.
Let \(\bar{\mathbf{X}} = S(\mathbf{X})\) 
represent the smoothed version of the raw time series 
 \(\mathbf{X}\), where $S$ is a  domain-specific smoothing function.
The forecasting task focuses on a  subset of electrical RTU signals indexed by  $E$,  and aims to predict the smoothed measurements  over a forecast  horizon 
$h$,  expressed as:
\(\bar{\mathbf{X}}[t:t+h, :E] \). 
Due to strict data export limitations imposed by system   security requirements,  we  work with  two complementary datasets:
\begin{itemize}
    \item A downsampled dataset with a 1-minute resolution ("1Min"), covering the full PSH system.
    \item A high-resolution dataset at 1-second resolution ("1Sec"),  limited to a subset of components due to bandwidth constraints.
\end{itemize}
To harmonize signals with varying native sampling rates, the raw downsampled time series are pre-processed  using simple moving averages within the high-security control environment. For  each sensor $j$, the effective sampling frequency  $S_j$  is used to compute the smoothed signal as:
\begin{equation}
    \bar{\mathbf{X}}[i,j] = S(\mathbf{X}[i,j]) =  \frac{1}{S_j} \sum_{\tau=1}^{S_j} s_\tau^j
\end{equation}
This preprocessing step ensures temporal consistency across heterogeneous sensor streams and mitigates high-frequency noise, improving model robustness and predictive accuracy.

\paragraph{Node Embeddings}
To encode short-term temporal dependencies at each sensor node, we use  two distinct \textit{Gated Recurrent Unit} (GRU) networks  as node-type-specific embedding functions, denoted by:
\(\text{Embed}_{\phi(v)}(\mathbf{S}[t-w:t]) \). 
Here, \( w \) is the window length, and \( \mathbf{S}_v[t-w:t] \) is the input slice assigned to node \( v \). The two GRUs use separate parameter sets for electrical and hydraulic node types and generate compact latent representations \( \mathbf{h}_v^{0} \in \mathbb{R}^{d_{\text{emb}}} \), computed as: 

\[
\mathbf{h}_v^{0} = 
\begin{cases} 
    \text{GRU}_{\text{el}}\big(W_{\text{emb}}^{\text{el}}\mathbf{x}_v^{t-w:t}\big), & v \in V_{\text{el}} \\
    \text{GRU}_{\text{hyd}}\big(W_{\text{emb}}^{\text{hyd}}\mathbf{x}_v^{t-w:t}\big), & v \in V_{\text{hyd}}
\end{cases}
\]

To incorporate external context and control inputs, we enrich the node representations  by appending relevant scalar features:
\[
\mathbf{h}_v^{t} = \text{Concat}\left(\mathbf{h}_v^{t}, \mathbf{U}_v^{t}\right),   for \quad v \in V_{\text{el}},
\]
where $\mathbf{U}_v^{t}$ includes control commands and sinusoidal time   encodings  (e.g. time of day,  day of  week). This enriched input   captures   both local  temporal patterns and global operational  context, which is crucial for accurate graph-based forecasting .

\textbf{Model Configuration \& Training:}
All  experiments were conducted 
using PyTorch 2.0 with CUDA 11.8 on an NVIDIA A100. 
For the \textit{1Min case study},  the model is configured with a lookback  window of $w=24$ minutes
and a prediction horizon of \(h=1\) minute, 
corresponding to forecasting the system’s average state one  minute  ahead.
For the \textit{1Sec case study}, which exhibits  step-wise dynamics (see  Figure \ref{fig:time_series_overview}),   the lookback window is extended to to (\(w=64\) seconds) to better capture fine-grained  temporal patterns. In this case, the model performs a two-step-ahead forecast (\(h=2\) seconds) to more accurately reflect short-term transitions.
The model architecture consists of three layers of heterogeneous graph attention, 
each followed by a leaky-ReLU activation function.  
This choice consistently  outperformed both ReLU and sigmoid activations in our empirical evaluations.
The design is tailored to the operational characteristics of the Swiss railway system,
which follows  a half-hourly  periodic timetable.
By aligning the input window with this schedule, the model  mitigates residual effects from prior intervals, thereby improving forecast consistency under structured  operating conditions.
Training is performed using the AdamW optimizer for 50 epochs or until convergence.
The best  model checkpoint is selected based on  validation  performance.

\begin{table}[h]
  \centering
\caption{Overview of trainable parameters in the neural network models for the \textit{1Min} and \textit{1Sec case studies}.}

\begin{tabular}{lll|lll}
\toprule
& & &  \textbf{\# Parameter} \\
\textbf{Model} &  &  & \textit{1Sec case study}  &  \textit{1Sec case study}  \\ \midrule
          LSTM*     & \cmark &                      &  3.17M   & 3.16M     \\
          LSTM & \cmark & \cmark  & 3.30M & 3.46M  \\     
          1D-CNN*    & \cmark &                      &  44k  & 18k       \\
          1D-CNN& \cmark & \cmark  & 4.3M & 288k  \\
          StemGNN   & \cmark & \cmark    &  3.38M & 544k                                     \\ \midrule
          StemGNN  & \cmark &   & 3.37M   & 514k \\
          GAT      & \cmark &           & 27k   & 179k \\ \bottomrule
          HGAT-small         & \cmark & \cmark       & -    &   91k           \\ 
          HGAT & \cmark & \cmark  & 100k  & 1.24M \\
\end{tabular}
\label{table:model_sizes}
\end{table}

\subsection{Baseline Models   for Evaluation }
\label{sec:baselines}

We evaluate   our approach against several widely  used data-driven 
baseline models for time-series forecasting in power systems , including \textit{1D Convolutional Neural Networks} and
\textit{Recurrent Neural Networks},
particularly   the Long Short-Term Memory (LSTM) model.
To establish a strong graph-based reference, we also   include a GNN model that uses the same architecture as our proposed  HGAT   
but without modeling   heterogeneous node and edge types  --
providing    a more meaningful baseline than a generic, off-the-shelf GNN.
Additionally, we assess the 
\textit{Spectral Temporal Graph Neural Network} (StemGNN), a state-of-the-art model specifically 
developed   for multivariate time-series forecasting.
We also  include a simple  persistence baseline,
which repeats the most recent observation; while naive, it  performs surprisingly well on the high-resolution \textit{1Sec case study} due to low   short-term variability .
To systematically  assess the  impact of   different inductive biases, we investigate the following aspects :

\begin{itemize}
    \item Role  of relational graph structure:
    We compare models that rely on the  explicitly defined   as sensor graph (as shown in Figure \ref{fig:graphs}) used by GAT and HGAT,
    against models like  StemGNN that infer relational structures directly from data .
    \item Signal-level vs.  system-level    training: We investigate how structuring the forecasting task at the individual sensor level, as opposed to modeling the system as a whole, influences training performance. Grouping signals by sensor enables   stochastic gradient descent to operate across   both temporal and spatial (sensor)  dimensions, encouraging   the learning of more localized and generalizable patterns . In contrast,  training the full system aggregates  all signals into a single input, limiting  stochastic variation to the temporal axis and potentially reducing model robustness . 
    We refer to  the signal-level variants of LSTM and 1D-CNN as LSTM* and 1D-CNN*, respectively .
    \item Derivative vs. absolute value forecasting: We compare models trained to  predict first-order forward differences  with those forecasting absolute sensor values. Derivative prediction explicitly models  system dynamics  and temporal change, which can improve sensitivity to operational variations. In contrast, directly forecasting absolute values bypasses the need for integration but ignores underlying system inertia, potentially leading to less accurate dynamic modeling.
\end{itemize}

\section{Results}

\label{sec:results}

This section presents  the numerical evaluation of the proposed \textit{Heterogeneous Graph ATtenton network} (HGAT) 
and compares its performance  to a range of baseline models.
Model performance is assessed using
normalized root mean square error (NRMSE) and 
normalized mean absolute error (NMAE).
In addition to overall performance comparisons, we analyze the impact of inductive biases introduced   in Section \ref{sec:baselines}, including  the role of relational structure and signal-level training.
To further assess the  contribution of multi-domain data, we conduct an ablation study by removing hydraulic signal inputs  from the model.
As noted earlier, 
models trained solely at the signal  level are  denoted  with an asterisk (LSTM* and 1D-CNN*).
All experiments are conducted  on the pumped-storage hydropower plant case studies
described  in Section \ref{sec:case_study}.

\begin{figure}[h]
\centering

\begin{subfigure}{\linewidth}
    \centering
    \includegraphics[width=\linewidth]{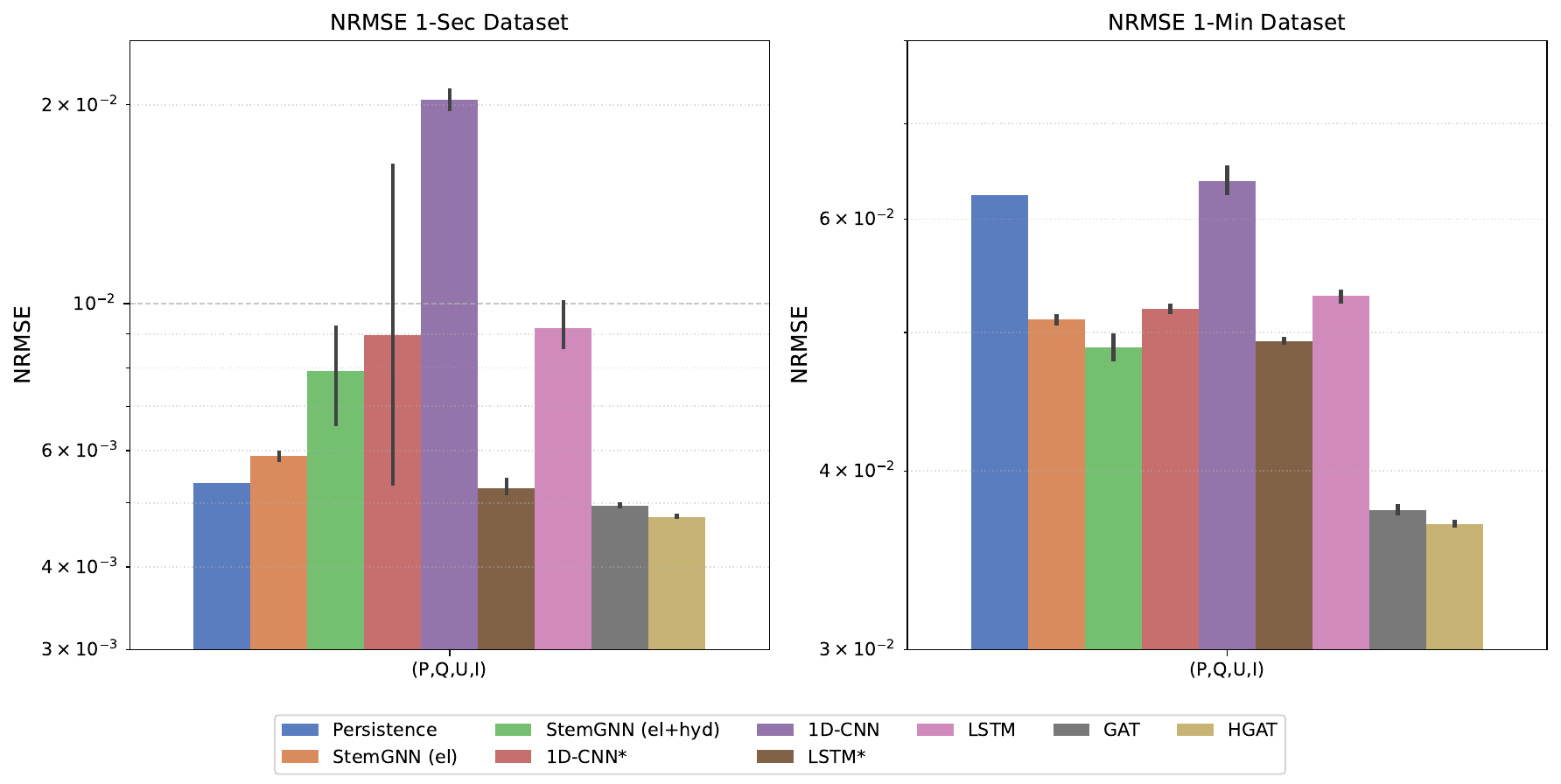}
    \caption{Aggregated over active power (P), reactive power (Q), voltage magnitude (U), and current magnitude (I).}
    \label{fig:results_rmse_top}
\end{subfigure}

\vspace{1em} % optional vertical spacing

\begin{subfigure}{\linewidth}
    \centering
    \includegraphics[width=\linewidth]{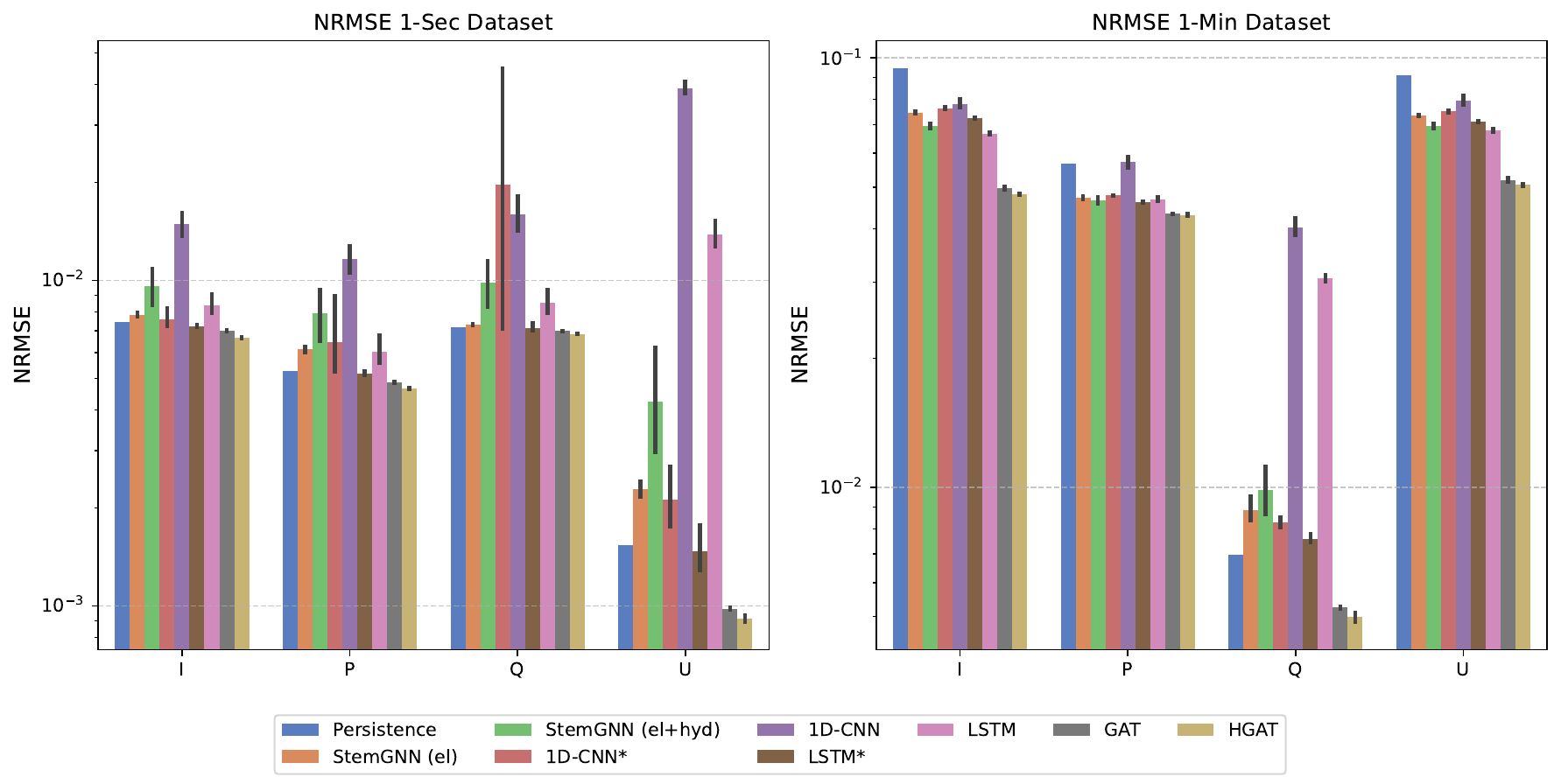}
    \caption{Per electrical state variable (P, Q, U, I).}
    \label{fig:results_rmse_bottom}
\end{subfigure}

\caption{Evaluation of the state forecasting error in terms of NRMSE on the \textit{1Min} and \textit{1Sec} case study.}
\label{fig:results_rmse_summarized}
\end{figure}

We begin by comparing 
HGAT to  the baseline models introduced 
in Section \ref{sec:baselines}.
As shown in Figure~\ref{fig:results_rmse_summarized}, 
 HGAT  consistently achieves the best 
 performance across all electrical state variables  
-- active power (P), 
reactive power (Q), 
voltage magnitude (U), 
and current magnitude (I) --
in terms of NRMSE, for both the \textit{1Min} and \textit{1Sec case studies}.
Average NRMSE and NMAE scores  for the  two case studies
are summarized in 
in Tables \ref{tab:results_1min}
and  \ref{tab:results_1sec}.
In the \textit{1Min} case study, HGAT  outperforms  all baselines in both metrics.
Specifically, in terms of NRMSE, HGAT achieves improvements over
LSTM (by 30.8\%), LSTM* (by 25.5\%), 1D-CNN (by 42.3\%), 1D-CNN* (by 29,2\%),
and StemGNN (el+hyd) (by 24,8\%).
In the \textit{1Sec} case study, the persistence baseline 
performs  strongly in terms of NMAE due to the step-wise behavior   of the signals, 
as illustrated in Figure \ref{fig:time_series_overview}.
While  none of models outperforms the persistence baseline in NMAE, the proposed
HGAT model  demonstrates clear advantages in NRMSE, especially in capturing rapid  changes in system state, achieving an improvement of 11.1\% .
Figure
\ref{fig:example_forecasts} highlights selected examples where HGAT provides  more responsive and accurate forecasts compared to the persistence baseline.
Overall,  results from  the \textit{1Sec case study} demonstrate  that  HGAT  significantly outperforms all other baselines in terms of NRMSE,
with relative improvements of:
(48.2\% under) LSTM, (9.3\% under) LSTM*, 
(76.6\% under) 1D-CNN, 
(46.8\% under) 1D-CNN*,
and (39.7\% under) StemGNN (el+hyd) .

In the GAT baseline,  hydraulic sensor data 
is excluded by removing all heterogeneous edges 
from the message-passing graph. This setup  
allows us to isolate and assess the contribution  
of fusing electrical and hydraulic sensor data 
in forecasting the states of the electrical subsystem.
Comparing GAT and HGAT in Tables \ref{tab:results_1min}
and  \ref{tab:results_1sec}
reveals that incorporating hydraulic information 
consistently improves performance 
across all state variables, as measured by both NMAE and NRMSE.
Specifically, HGAT outperforms GAT in terms of NRMSE by 3.8\% in the \textit{1Sec case study} and by 2.2\% in the \textit{1Min case study}.
Notably, this improvement is only observed  
when the model explicitly  incorporates  the heterogeneous graph structure.
In contrast,
augmenting other baseline models 
with hydraulic information tends to degrade their  performance.  For instance,
StemGNN (el) outperforms  StemGNN (el+hyd) 
by 25.5\% on the \textit{1Sec case study}, while the opposite is true in the 
\textit{1Min case study}, where the extended version improves performance by 4.4\% .
These findings highlight  the effectiveness of HGAT in leveraging heterogeneous graph structures to capture cross-domain interactions. They also indicate that while models like StemGNN can implicitly learn  inter-domain relationships  in smaller graphs
-- as demonstrated in \cite{theilerGraphNeuralNetworks2024} -- 
their performance degrades as the graph complexity  increases. Moreover, removing the  heterogeneous structure from  HGAT and HGNN, using the GAT and GNN model instead,
leads to  significantly higher performance variance across  retrainings on average (Tables \ref{tab:results_1min} and \ref{tab:results_1sec}), underlining  the stabilizing role of explicitly modeled heterogeneity.

Interestingly, the signal-wise trained models  LSTM* and 1D-CNN* outperform 
their global counterparts (LSTM and 1D-CNN)
that were trained on the full set of signals, 
despite not having access to the full signal state.
In Figure \ref{fig:methodology_overview}, this approach corresponds to selecting LSTM or 1D-CNN as time encoder network and directly mapping from the latent time encoding to the state forecast, bypassing the heterogeneous graph processing layers.
1D-CNN* outperforms 1D-CNN by 56.0\% in the \textit{1Sec case study} 
and by 18.7\% in the \textit{1Min case study}.  Similarly, LSTM* outperfrorms LSTM by 42.8\% and  7.0\% in the \textit{1Sec} 
and \textit{1Min case studies}, respectively. The largest performance gaps are observed   in reactive power prediction (1Min) and voltage magnitude prediction (1Sec), as shown in Figure \ref{fig:results_rmse_summarized}.
These results suggest  that LSTM and 1D-CNN struggle to capture  meaningful inter-sensor relationships  when trained on  the full  multivariate signal 
and are more prone to  overfitting to global  state trends.

The proposed HGAT framework maintains  
 competitive performance even in its lightweight version  (HGAT-small), which uses significantly fewer trainable parameters. Scaling up  to the full HGAT
 model yields an additional  1.94 \% in NRMSE. 
A breakdown  of model sizes is provided in Table
\ref{table:model_sizes}.
Additional  evaluation in the Appendix (Figures \ref{fig:results_detailed_1min_case_study} and \ref{fig:results_detailed_1sec_case_study}) provides  
per-node performance analysis, confirming that HGAT improvements are consistently distributed across sensor locations,
ensuring that no node  experiences a performance  drop when hydraulic sensor information is included.
This highlights the method’s robustness and generalizability  across hybrid energy systems.

A comparison of results across the two case studies reveals that  baseline model  performance is sensitive to sampling rate.
Figure \ref{fig:results_rmse_summarized} shows that StemGNN outperforms 
the simpler LSTM and 1D-CNN baselines 
for reactive power forecasting in the \textit{1Sec case study}, 
but this trend reverses in the \textit{1Min case study}.
In contrast,  voltage  magnitude prediction remains relatively stable  across models 
in the \textit{1Min case study} but varies  significantly in the \textit{1Sec case study}.

To further evaluate the effectiveness of our architectural choices, we conduct an ablation study on the components proposed in the HGAT framework (see Figure \ref{fig:methodology_overview}). Tables \ref{tab:results_1min} and  \ref{tab:results_1sec} compare the use of first-order Euler integration (HGAT) with the direct direct  prediction (HGAT-D), as well as graph attention (GATv2) with standard graph convolution in the homogeneous message passing function (HGNN). Our results demonstrate that combining first-order Euler integration with HGAT improves performance in the \textit{1Min case study} by 1.15\% in NRMSE across all target  variables. In the \textit{1Sec case study},  overall 
performance remains  largely unchanged, except for a minor degradation in active power forecasting. Comparing HGAT and HGNN  yields  a 16.3\% improvement in NRMSE in the \textit{1Min case study}, while the \textit{1Sec case study} shows a negligible decrease (0.58\%).
These results likely reflect the nature of the \textit{1Sec case study}, where signals are often stationary for extended durations before exhibiting sharp state transitions, making the core task one of change-point detection rather than continuous forecasting, as the next data point's value is given in the input window with high probability and thus is seen by the model.

In summary, the results  clearly show   that   the best overall performance is achieved by modeling signals
at the signal level  using a recurrent encoder, 
followed by message passing  over  a heterogeneous graph, 
and finally predicting first-order forward  differences. 
This pipeline -- visualized  in Figure \ref{fig:methodology_overview} -- proves to be both effective and robust across different sensor modalities and sampling rates. Moreover, HGAT consistently shows lower variance across retrainings compared to all baselines, further underscoring its stability and suitability for practical deployment in hybrid power systems.

\begin{table*}[htbp]
\centering
\caption{Average (normalized) model performance averaged across nodes for the \textit{1Min case study}.
We indicate whether the PSH network diagrams were translated into a processable graph for the computation (Network Diagram)
and epmhasize if hydraulic (hyd) or electric (el) information was used for training.
The results are scaled by a factor of $10^2$ to enhance readability and are rounded to two decimal places.}
\makebox[\textwidth][c]{%
\begin{tabular}{lllllllllll}
\toprule

&&& \multicolumn{2}{l}{P} & \multicolumn{2}{l}{Q}  & \multicolumn{2}{l}{U}  & \multicolumn{2}{l}{I} \\
\cmidrule(l){4-5} \cmidrule(l){6-7} \cmidrule(l){8-9} \cmidrule(l){10-11} 
Ablation & el & hyd & NMAE $\downarrow$ & NRMSE $\downarrow$ & NMAE $\downarrow$ & NRMSE $\downarrow$  & NMAE $\downarrow$ & NRMSE $\downarrow$  & NMAE $\downarrow$& NRMSE $\downarrow$     \\
\midrule
1D-CNN*& \cmark &  & \text{\small 3.30} \text{\tiny ±0.03} & \text{\small 4.92} \text{\tiny ±0.02} & \text{\small 0.30} \text{\tiny ±0.05} & \text{\small 0.98} \text{\tiny ±0.04} & \text{\small 4.92} \text{\tiny ±0.08} & \text{\small 7.63} \text{\tiny ±0.06} & \text{\small 4.72} \text{\tiny ±0.05} & \text{\small 7.97} \text{\tiny ±0.05} \\
1D-CNN & \cmark & \cmark& \text{\small 4.30} \text{\tiny ±0.18} & \text{\small 5.80} \text{\tiny ±0.20} & \text{\small 1.80} \text{\tiny ±0.21} & \text{\small 5.63} \text{\tiny ±0.31} & \text{\small 5.66} \text{\tiny ±0.25} & \text{\small 8.06} \text{\tiny ±0.28} & \text{\small 5.35} \text{\tiny ±0.19} & \text{\small 8.11} \text{\tiny ±0.16} \\
LSTM* & \cmark & & \text{\small 3.17} \text{\tiny ±0.02} & \text{\small 4.74} \text{\tiny ±0.01} & \text{\small 0.30} \text{\tiny ±0.01} & \text{\small 0.90} \text{\tiny ±0.02} & \text{\small 4.58} \text{\tiny ±0.06} & \text{\small 7.23} \text{\tiny ±0.04} & \text{\small 4.38} \text{\tiny ±0.05} & \text{\small 7.59} \text{\tiny ±0.04} \\
LSTM & \cmark & \cmark& \text{\small 3.39} \text{\tiny ±0.07} & \text{\small 4.85} \text{\tiny ±0.06} & \text{\small 1.04} \text{\tiny ±0.10} & \text{\small 4.58} \text{\tiny ±0.07} & \text{\small 4.68} \text{\tiny ±0.10} & \text{\small 6.87} \text{\tiny ±0.08} & \text{\small 4.36} \text{\tiny ±0.08} & \text{\small 6.89} \text{\tiny ±0.06} \\
StemGNN & \cmark &  & \text{\small 3.27} \text{\tiny ±0.11} & \text{\small 4.77} \text{\tiny ±0.11} & \text{\small 0.42} \text{\tiny ±0.16} & \text{\small 1.11} \text{\tiny ±0.15} & \text{\small 4.60} \text{\tiny ±0.13} & \text{\small 7.03} \text{\tiny ±0.10} & \text{\small 4.32} \text{\tiny ±0.13} & \text{\small 7.17} \text{\tiny ±0.13} \\
StemGNN & \cmark & \cmark& \text{\small 3.33} \text{\tiny ±0.06} & \text{\small 4.84} \text{\tiny ±0.05} & \text{\small 0.32} \text{\tiny ±0.04} & \text{\small 1.02} \text{\tiny ±0.06} & \text{\small 4.79} \text{\tiny ±0.04} & \text{\small 7.45} \text{\tiny ±0.03} & \text{\small 4.59} \text{\tiny ±0.07} & \text{\small 7.77} \text{\tiny ±0.05} \\
GNN & \cmark & & \text{\small 3.36} \text{\tiny ±0.11} & \text{\small 5.00} \text{\tiny ±0.11} & \text{\small 0.74} \text{\tiny ±0.14} & \text{\small 4.17} \text{\tiny ±0.07} & \text{\small 3.92} \text{\tiny ±0.11} & \text{\small 6.06} \text{\tiny ±0.12} & \text{\small 3.60} \text{\tiny ±0.14} & \text{\small 6.24} \text{\tiny ±0.17} \\ 
HGNN & \cmark & & \text{\small 3.18} \text{\tiny ±0.02} & \text{\small 4.69} \text{\tiny ±0.03} & \text{\small 0.48} \text{\tiny ±0.04} & \text{\small 3.47} \text{\tiny ±0.29} & \text{\small 3.74} \text{\tiny ±0.09} & \text{\small 5.72} \text{\tiny ±0.07} & \text{\small 3.32} \text{\tiny ±0.07} & \text{\small 5.35} \text{\tiny ±0.08} \\ \midrule
Persistence&  &  & \text{\small 3.81}  & \text{\small 5.88}  & \text{\small \underline{0.09}} & \text{\small 1.01} & \text{\small 5.62}  & \text{\small 9.28}  & \text{\small 5.35}  & \text{\small 9.93}  \\\midrule
GAT & \cmark & \cmark& \text{\small 2.97} \text{\tiny ±0.01} & \text{\small 4.49} \text{\tiny ±0.01} & \text{\small \underline{0.09}} \text{\tiny ±0.00} & \text{\small 0.71} \text{\tiny ±0.00} & \text{\small 3.52} \text{\tiny ±0.04} & \text{\small 5.38} \text{\tiny ±0.06} & \text{\small \underline{3.13}} \text{\tiny ±0.03} & \text{\small 5.09} \text{\tiny ±0.05} \\ 
HGAT-D & \cmark & \cmark& \text{\small \underline{2.96}} \text{\tiny ±0.01} & \text{\small \underline{4.48}} \text{\tiny ±0.02} & \text{\small 0.12} \text{\tiny ±0.01} & \text{\small 0.72} \text{\tiny ±0.01} & \text{\small \underline{3.40}} \text{\tiny ±0.03} & \text{\small \underline{5.28}} \text{\tiny ±0.05} & \text{\small \textbf{3.00}} \text{\tiny ±0.02} & \text{\small \underline{4.98}} \text{\tiny ±0.05} \\
HGAT-small & \cmark & \cmark& \text{\small 2.97} \text{\tiny ±0.02} & \text{\small 4.49} \text{\tiny ±0.02} & \text{\small \underline{0.09}} \text{\tiny ±0.00} & \text{\small \underline{0.70}} \text{\tiny ±0.02} & \text{\small 3.51} \text{\tiny ±0.05} & \text{\small 5.35} \text{\tiny ±0.05} & \text{\small \underline{3.13}} \text{\tiny ±0.06} & \text{\small 5.07} \text{\tiny ±0.06} \\
HGAT & \cmark & \cmark& \text{\small \textbf{2.93}} \text{\tiny ±0.02} & \text{\small \textbf{4.45}} \text{\tiny ±0.03} & \text{\small \textbf{0.08}} \text{\tiny ±0.00} & \text{\small \textbf{0.68}} \text{\tiny ±0.02} & \text{\small \textbf{3.39}} \text{\tiny ±0.03} & \text{\small \textbf{5.24}} \text{\tiny ±0.03} & \text{\small \textbf{3.00}} \text{\tiny ±0.01} & \text{\small \textbf{4.95}} \text{\tiny ±0.01} \\ 
\bottomrule
\end{tabular}
}
\label{tab:results_1min}
\end{table*}

\begin{table*}[htbp]
  \centering
  \caption{Average (normalized) model performance across sensor nodes in the \textit{1Sec case study}.
  We indicate whether the physical network schematics were incorporated as processable graph structure  (Network Diagram)
  and highlight  whether electrical (el)  and/or hydraulic (hyd) data  were used during  training.
  All  results are scaled by a factor of $10^2$ for  readability and rounded to two decimal places.}

\makebox[\textwidth][c]{%
\begin{tabular}{lllllllllll}
\toprule

& & & \multicolumn{2}{l}{P} & \multicolumn{2}{l}{Q}  & \multicolumn{2}{l}{U}  & \multicolumn{2}{l}{I} \\
\cmidrule(l){4-5} \cmidrule(l){6-7} \cmidrule(l){8-9} \cmidrule(l){10-11} 
Ablation & el & hyd & NMAE $\downarrow$ & NRMSE $\downarrow$ & NMAE $\downarrow$ & NRMSE $\downarrow$  & NMAE $\downarrow$ & NRMSE $\downarrow$  & NMAE $\downarrow$& NRMSE $\downarrow$     \\
\midrule

1D-CNN* & \cmark & & \text{\small 0.45} \text{\tiny ±0.10} & \text{\small 0.76} \text{\tiny ±0.07} & \text{\small 0.42} \text{\tiny ±0.29} & \text{\small 0.65} \text{\tiny ±0.28} & \text{\small 1.40} \text{\tiny ±2.50} & \text{\small 1.97} \text{\tiny ±2.81} & \text{\small 0.12} \text{\tiny ±0.05} & \text{\small 0.21} \text{\tiny ±0.06} \\
1D-CNN & \cmark & \cmark& \text{\small 0.95} \text{\tiny ±0.07} & \text{\small 1.49} \text{\tiny ±0.15} & \text{\small 0.73} \text{\tiny ±0.07} & \text{\small 1.17} \text{\tiny ±0.15} & \text{\small 1.01} \text{\tiny ±0.19} & \text{\small 1.60} \text{\tiny ±0.24} & \text{\small 1.16} \text{\tiny ±0.08} & \text{\small 3.89} \text{\tiny ±0.21} \\
LSTM* & \cmark & & \text{\small 0.41} \text{\tiny ±0.04} & \text{\small 0.72} \text{\tiny ±0.01} & \text{\small 0.29} \text{\tiny ±0.01} & \text{\small 0.52} \text{\tiny ±0.01} & \text{\small 0.28} \text{\tiny ±0.03} & \text{\small 0.71} \text{\tiny ±0.03} & \text{\small 0.05} \text{\tiny ±0.02} & \text{\small 0.15} \text{\tiny ±0.03} \\
LSTM & \cmark & \cmark& \text{\small 0.55} \text{\tiny ±0.06} & \text{\small 0.84} \text{\tiny ±0.07} & \text{\small 0.41} \text{\tiny ±0.06} & \text{\small 0.60} \text{\tiny ±0.08} & \text{\small 0.49} \text{\tiny ±0.10} & \text{\small 0.85} \text{\tiny ±0.09} & \text{\small 0.43} \text{\tiny ±0.06} & \text{\small 1.38} \text{\tiny ±0.16} \\ 
StemGNN & \cmark & & \text{\small 0.47} \text{\tiny ±0.01} & \text{\small 0.78} \text{\tiny ±0.01} & \text{\small 0.39} \text{\tiny ±0.01} & \text{\small 0.62} \text{\tiny ±0.02} & \text{\small 0.31} \text{\tiny ±0.02} & \text{\small 0.73} \text{\tiny ±0.01} & \text{\small 0.16} \text{\tiny ±0.02} & \text{\small 0.23} \text{\tiny ±0.02} \\
StemGNN & \cmark & \cmark& \text{\small 0.55} \text{\tiny ±0.07} & \text{\small 0.96} \text{\tiny ±0.15} & \text{\small 0.46} \text{\tiny ±0.07} & \text{\small 0.79} \text{\tiny ±0.18} & \text{\small 0.46} \text{\tiny ±0.10} & \text{\small 0.99} \text{\tiny ±0.21} & \text{\small 0.24} \text{\tiny ±0.07} & \text{\small 0.42} \text{\tiny ±0.22} \\
GNN & \cmark & & \text{\small 0.40} \text{\tiny ±0.01} & \text{\small 0.68} \text{\tiny ±0.00} & \text{\small 0.28} \text{\tiny ±0.00} & \text{\small \underline{0.48}} \text{\tiny ±0.00} & \text{\small 0.27} \text{\tiny ±0.01} & \text{\small \underline{0.69}} \text{\tiny ±0.00} & \text{\small \underline{0.02}} \text{\tiny ±0.00} & \text{\small \underline{0.10}} \text{\tiny ±0.00} \\
HGNN & \cmark & \cmark& \text{\small 0.39} \text{\tiny ±0.01} & \text{\small \underline{0.66}} \text{\tiny ±0.00} & \text{\small \underline{0.26}} \text{\tiny ±0.00} & \text{\small \textbf{0.46}} \text{\tiny ±0.00} & \text{\small 0.28} \text{\tiny ±0.00} & \text{\small \underline{0.69}} \text{\tiny ±0.00} & \text{\small \underline{0.02}} \text{\tiny ±0.00} & \text{\small \textbf{0.09}} \text{\tiny ±0.00} \\\midrule
Persistence &  & & \text{\small \textbf{0.28}} & \text{\small 0.75} & \text{\small \textbf{0.20}} & \text{\small 0.53} & \text{\small \textbf{0.17}} & \text{\small 0.72}  & \text{\small \textbf{0.01}} & \text{\small 0.15} \\ \midrule
GAT & \cmark & & \text{\small 0.41} \text{\tiny ±0.01} & \text{\small 0.70} \text{\tiny ±0.00} & \text{\small 0.28} \text{\tiny ±0.00} & \text{\small 0.49} \text{\tiny ±0.00} & \text{\small \underline{0.26}} \text{\tiny ±0.00} & \text{\small 0.70} \text{\tiny ±0.00} & \text{\small \underline{0.02}} \text{\tiny ±0.00} & \text{\small \underline{0.10}} \text{\tiny ±0.00} \\
HGAT-D & \cmark & \cmark& \text{\small \underline{0.37}} \text{\tiny ±0.00} & \text{\small \textbf{0.65}} \text{\tiny ±0.00} & \text{\small \underline{0.26}} \text{\tiny ±0.00} & \text{\small \textbf{0.46}} \text{\tiny ±0.00} & \text{\small \underline{0.26}} \text{\tiny ±0.00} & \text{\small \textbf{0.68}} \text{\tiny ±0.00} & \text{\small \underline{0.02}} \text{\tiny ±0.00} & \text{\small \textbf{0.09}} \text{\tiny ±0.00} \\
HGAT &\cmark & \cmark& \text{\small 0.39} \text{\tiny ±0.00} & \text{\small 0.67} \text{\tiny ±0.00} & \text{\small 0.27} \text{\tiny ±0.00} & \text{\small \textbf{0.46}} \text{\tiny ±0.00} & \text{\small 0.27} \text{\tiny ±0.01} & \text{\small \textbf{0.68}} \text{\tiny ±0.00} & \text{\small \underline{0.02}} \text{\tiny ±0.00} & \text{\small \textbf{0.09}} \text{\tiny ±0.00} \\
\bottomrule
\end{tabular}
}
\label{tab:results_1sec}
\end{table*}

\clearpage

\section{Conclusions}
\label{sec:conclusions}

In this paper, we investigated how heterogeneous data sources in power systems,
recorded by different sensor types, 
spanning multiple physical domains, 
and operating at varying temporal resolutions 
can be leveraged for data-driven short-term state forecasting.
Such rich, multi-modal data is becoming increasingly available through modern energy management systems, offering new opportunities to improve forecasting accuracy.

We propose a novel \textit{Heterogeneous Graph Attention Network} , 
the first approach to apply attention-based message passing across 
domains in energy conversion and distribution systems.  
Our method effectively captures 
system dynamics
at varying temporal resolutions, enabling accurate short-term forecasting, 
while remaining robust to noisy data. 
Our work systematically examined  the impact of several inductive biases, including the role of the relational graph structure, 
the benefits of signal-level versus system-level training, and 
the effects of forecasting derivatives versus  absolute values.

Through two real-world case studies,
we demonstrated  that 
integrating measurements from both electrical and hydraulic subsystems, 
grouping signals by spatial sensor site
and predicting first-order forward differences
leads to the most  and accurate state forecasting in pumped-storage hydropower plants.  These improvements are realized only when inter-sensor relationships are explicitly modeled as a heterogeneous graph.

Unlike physics-based simulation, 
our approach requires only a coarse representation of 
 sensor placement and does not rely on detailed physical topology or labor-intensive calibration.
This makes it practical and scalable for deployment in real-world systems. Our results demonstrate  
that explicitly modeling cross-domain  interactions
-- especially relevant in increasingly complex systems with  
renewable and distributed energy components --  
yields substantially improved  forecasting performance  compared to conventional 
deep learning models like  1D-CNN or LSTM, which treat all signals uniformly.
 
Importantly, the proposed approach retains  interpretability by leveraging graph structures 
that reflect real-world system architecture. This stands in contrast to  other deep learning methods 
that rely on inferred topologies or
 operate without any structural prior, often leading to reduced interpretability and limited generalization.
Future work will explore dynamic topologies  by incorporating time-varying
system elements, such as switch and breaker  states, directly into the graph representation.
Additionally, integrating physics-informed loss functions, 
 for example those based on 
generator efficiency or power flow constraints, 
holds strong potential to further improve model fidelity, physical consistency,  and robustness.

\section*{Acknowledgments}
This research was funded by the Swiss Federal Office of Transport (FOT) under the project INtellIgenT maIntenance rAilway power sysTEms (INITIATE). The authors would like to thank FOT for the project coordination and Swiss Federal Railways (SBB) for providing the data for this research and the discussions on the research results and the paper.

\clearpage

\bibliographystyle{unsrtnat}  
\bibliography{2025_HeteroGNN}
% ---------------------------------------------------------------------------------

\newpage
\appendix

\section{Additional Experimental Results}

\begin{figure}[h]
\centering
%\fbox{\rule[-.5cm]{0cm}{5cm} \rule[-.5cm]{5cm}{0cm}}
\includegraphics[width=1\linewidth]{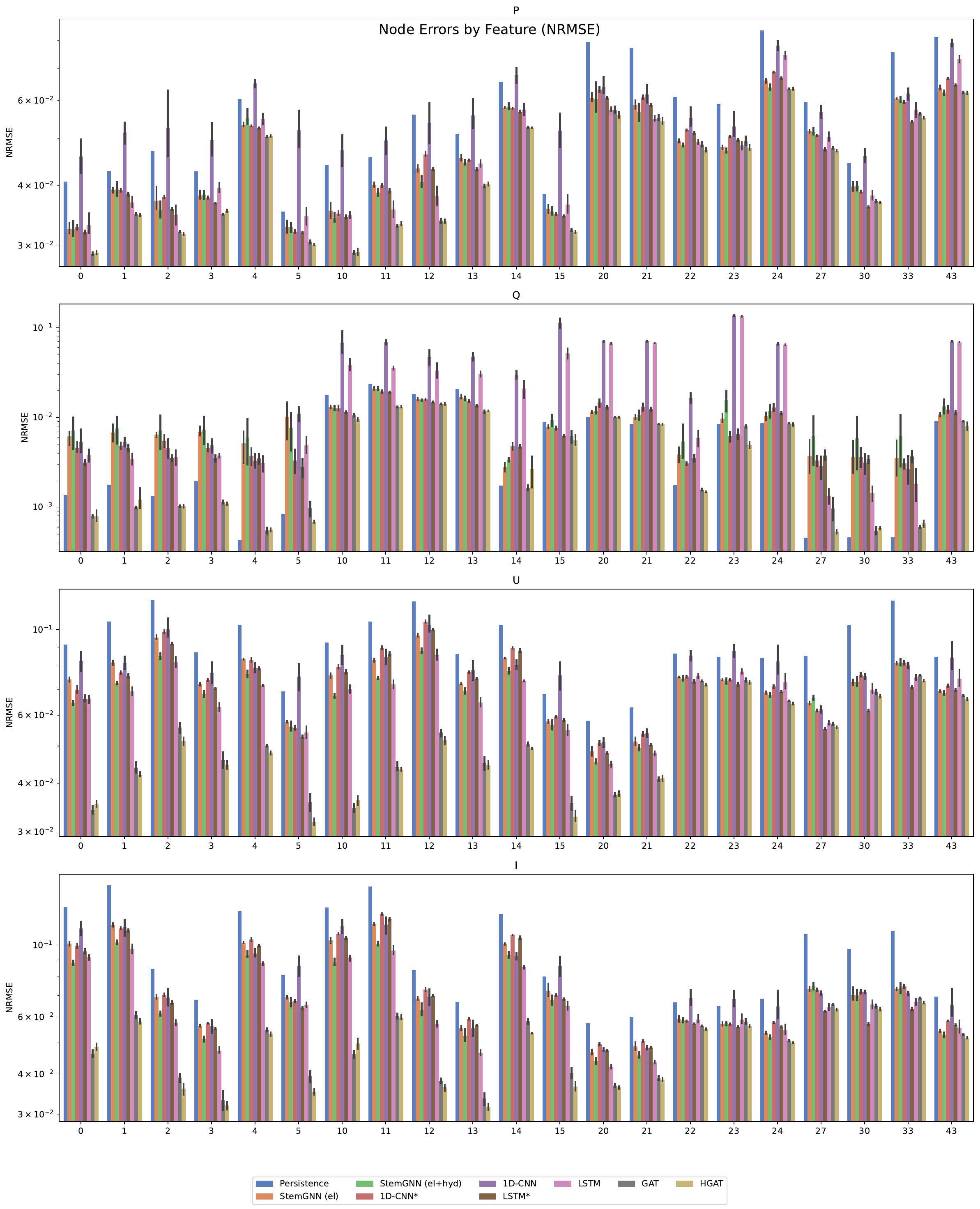}

\caption{\textit{1Sec case study}: detailed evaluation of the state forecsting error in terms of NRMSE by node and feature.}
\label{fig:results_detailed_1min_case_study}
\end{figure}

\begin{figure}[h]
\centering
%\fbox{\rule[-.5cm]{0cm}{5cm} \rule[-.5cm]{5cm}{0cm}}
\includegraphics[width=0.8\linewidth]{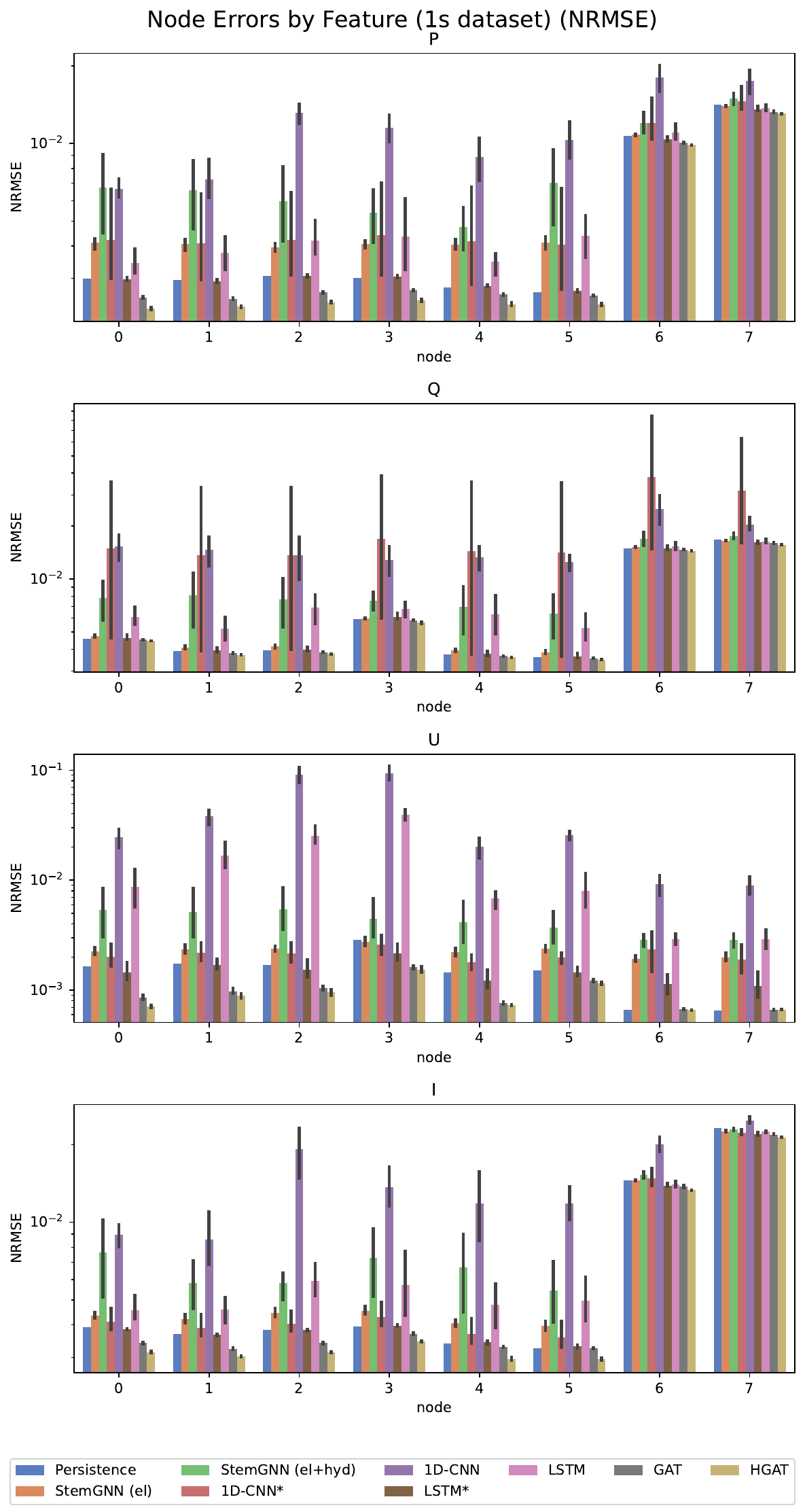}

\caption{\textit{1Sec case study}: detailed evaluation of the state forecsting error in terms of NRMSE by node and feature.}
\label{fig:results_detailed_1sec_case_study}
\end{figure}

\begin{figure}[h]
    \centering
    %\fbox{\rule[-.5cm]{0cm}{5cm} \rule[-.5cm]{5cm}{0cm}}
    % Syntax: \includegraphics[trim={left bottom right top},clip]{filename}
    % \includegraphics[trim={3cm 3cm 3cm 3cm}, width=1\linewidth]{Figures/example_days.pdf}
    \includegraphics[width=1\linewidth]{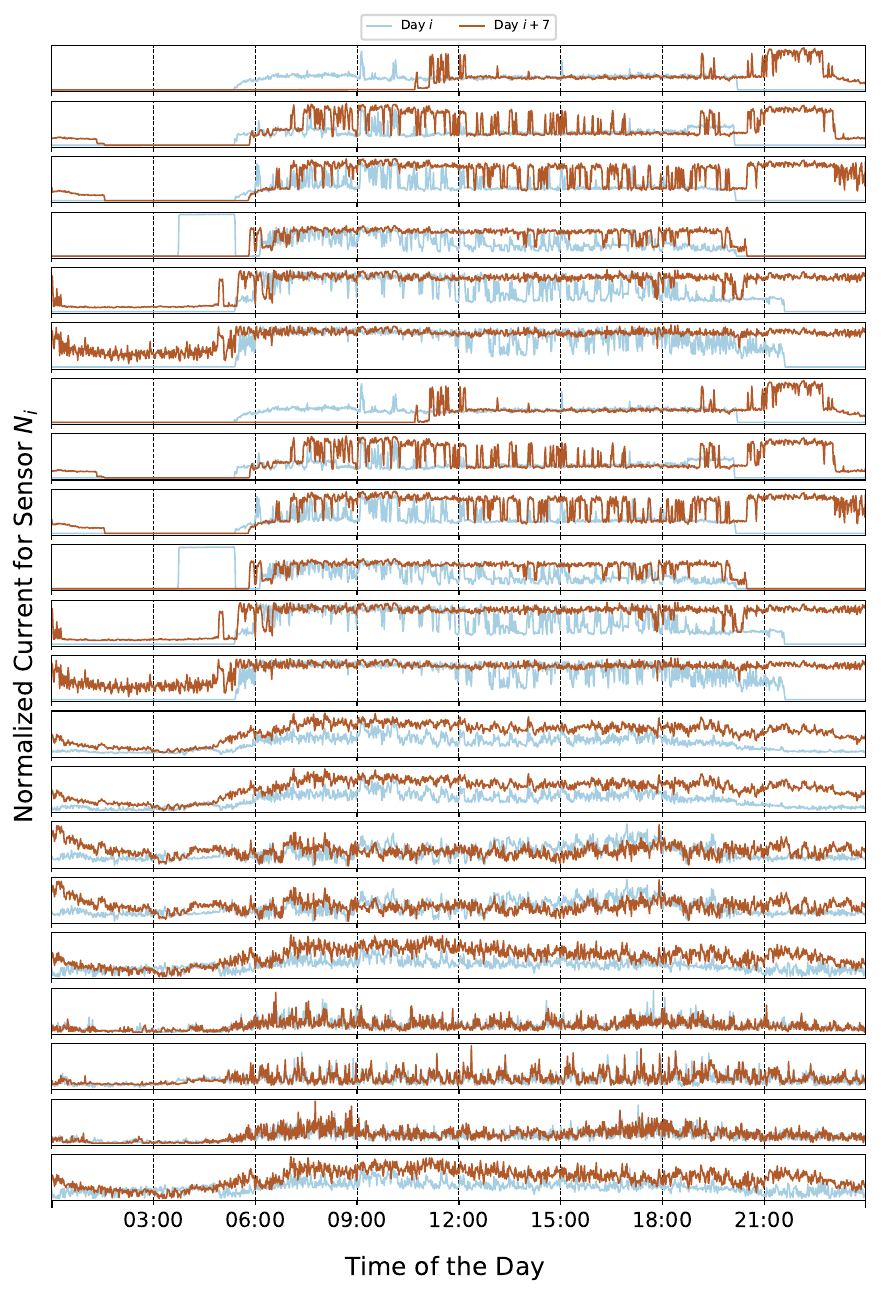}
  \caption{Segment of the \textit{1Sec case study}, displaying all normalized currents, 
  indicating the dynamic nature of the sensor measurements.  We show the same day of the week ($i$ and $i+7$) for two consecutive weeks.}
      \label{fig:zoomed_dataset}
\end{figure}

\begin{figure}[h]
    \centering
    %\fbox{\rule[-.5cm]{0cm}{5cm} \rule[-.5cm]{5cm}{0cm}}
    % Syntax: \includegraphics[trim={left bottom right top},clip]{filename}
    % \includegraphics[trim={3cm 3cm 3cm 3cm}, width=1\linewidth]{Figures/example_days.pdf}
    \includegraphics[width=1\linewidth]{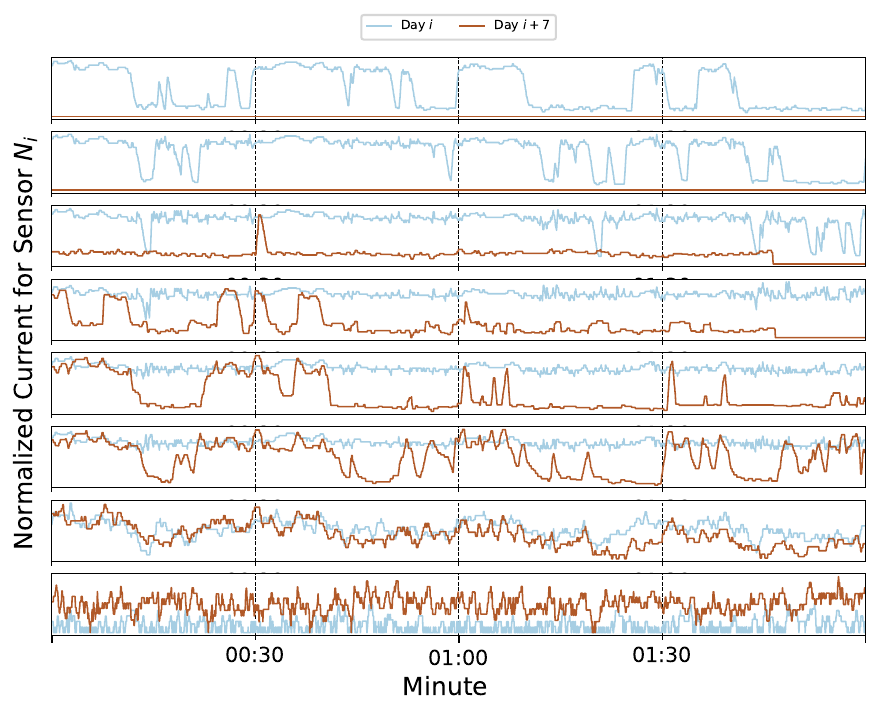}
  \caption{Segment of the \textit{1Sec case study}, displaying normalized currents, indicating the dynamic nature of the sensor measurements.  We show the same minutes of the week ($i$ and $i+7$) for two consecutive weeks.}
      \label{fig:zoomed_dataset_1sec}
\end{figure}

\begin{figure}[h]
    \centering
    %\fbox{\rule[-.5cm]{0cm}{5cm} \rule[-.5cm]{5cm}{0cm}}
    % Syntax: \includegraphics[trim={left bottom right top},clip]{filename}
    % \includegraphics[trim={3cm 3cm 3cm 3cm}, width=1\linewidth]{Figures/example_days.pdf}
    \includegraphics[width=1\linewidth]{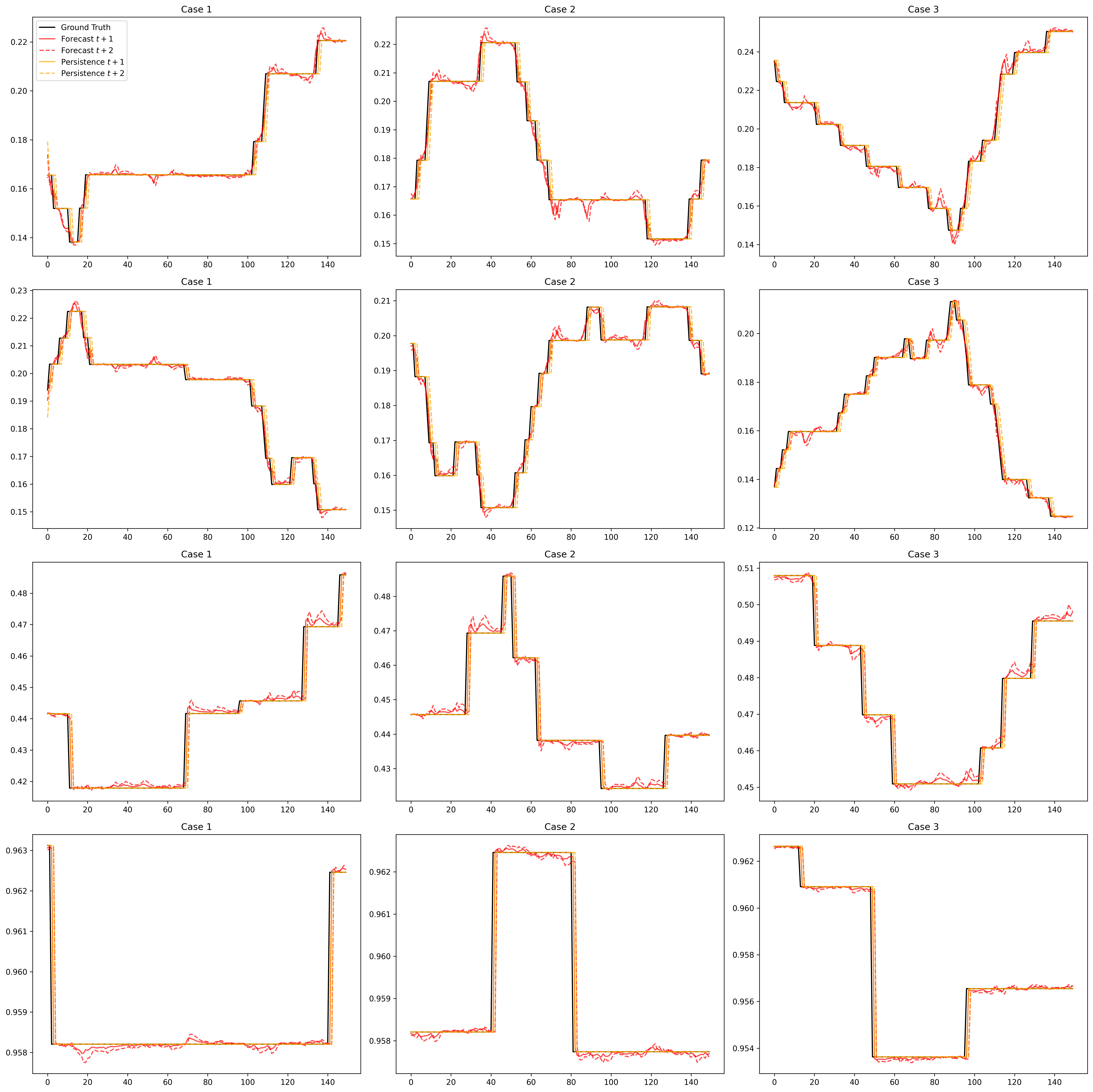}
  \caption{Forecast examples showing the performance of the HGAT model compared to the persistence baseline, the second-best performer on the 1Sec dataset. We forecast one ($t+1$) and two steps ahead ($t+2$).}
      \label{fig:example_forecasts}
\end{figure}

\end{document}